\documentclass{article}

\usepackage[T1]{fontenc}
\usepackage{microtype}
\usepackage{graphicx}
\usepackage{booktabs}
\usepackage{enumitem}
\usepackage{amsmath,amssymb,amsthm,mathtools}
\usepackage[colorlinks=true,allcolors=blue]{hyperref}
\usepackage[nameinlink,noabbrev]{cleveref}
\usepackage{placeins}
\usepackage[preprint]{icml2026}

\newtheorem{theorem}{Theorem}

\newcommand{\Comm}{\operatorname{Comm}}

\newcommand{\dd}{\mathrm{d}}

\icmltitlerunning{Discovering Physical Representation Languages}

\begin{document}
\twocolumn[
\icmltitle{Discovering Physical Representation Languages}
\begin{icmlauthorlist}
  \icmlauthor{Linzhe Zhang}{neu}
  \icmlauthor{Changming Xu}{neu}
\end{icmlauthorlist}
\icmlaffiliation{neu}{Graduate School, Northeastern University}
\icmlcorrespondingauthor{Linzhe Zhang}{cfmy007@gmail.com}
\icmlcorrespondingauthor{Changming Xu}{changmingxu@neuq.edu.cn}
\icmlkeywords{Scientific Machine Learning, Ontology Discovery, Representation Learning}
\vskip 0.3in
]
\printAffiliationsAndNotice{}

\begin{abstract}
Before a machine can discover a physical law, it must discover what its
measurements \emph{are}: which observations live on cells, which are intensive or
extensive, which sectors are dual, and which distinctions are merely gauge.  We
introduce \emph{physical representation-language discovery}, the problem of
recovering this hidden ontology directly from anonymous controlled experiments.
We give an identifiability theory and constructive polynomial-time procedure that
recovers a carrier and differential sequence, measurement types and orientation
twist, noninvertible refinement semantics, primal--dual Maxwell diagrams, and the
residual equivalences that no permitted experiment can break.  The theory turns
material nuisance into a commutant, uses refinement to separate quantities from
coordinates, and selects physics only after its representation has been recovered.
For a certified finite experiment family, we prove an end-to-end two-stage
measurement bound and a matching minimax rate in dimension, accuracy, and confidence.
Blind Maxwell experiments recover complete primal/relative-dual ontologies on
regular and unstructured carriers under jointly corrupted observations; an
independent unstructured RLC system demonstrates that the result is not specific to
Maxwell.  The framework scales to tens of thousands of cells per carrier,
while stress audits demonstrate robustness across severe physical
regimes---including non-Markovian memory, nonlinearities, nonlocality, and
complex constitutive hysteresis.
A public FDTD audit demonstrates the emergence of anonymous curl structure
from incomplete field data, while characterizing the informational prerequisites
for complete recovery.  The goal is to move scientific ML from learning laws in
a human-supplied language to discovering the language in which laws become
expressible, establishing exact theoretical limits on observational identifiability.
\end{abstract}

\section{Introduction}

Equation discovery assumes a state vector and a library of candidate laws
\citep{brunton2016sindy}; discrete physics learning normally assumes an exact
sequence or cell complex \citep{trask2022dec,degoes2016subdivision}.  Symmetry
methods have relaxed a neighboring assumption: AtlasD discovers local symmetries
\citep{bhat2025atlasd}, while Equivariance by Contrast (EbC) identifies
equivariant embeddings from action pairs \citep{schmidt2025equivariance}.
These methods still do not return the \emph{representation language} in which an
unknown physical law is expressed.

We define a representation language as a carrier face poset, ranks, straight/twisted
types, refinement transports, differential diagram, constitutive/dual relations,
and the unbroken gauge symmetries.  Our target is the intrinsic observational
equivalence class resolvable under a controlled experiment family, grounding physical
representation in operational distinguishability rather than arbitrary coordinate labels.

In summary, this work makes five primary contributions: (i) We characterize material-nuisance ambiguity as a
commutant and show how two probes collapse it to scale.  (ii) We prove that
noninvertible refinement distinguishes summation from measure-weighted averaging,
and that a separating fingerprint plus boundary recovers a carrier incidence
algebra.  (iii) We combine these primitives into blind primal/dual Maxwell recovery
and supply finite-sample, conditioning, noise, scale, and fair-baseline evidence.
(iv) We give an end-to-end quotient-space sample-complexity theorem with an explicit
two-stage reused-readout budget and matching minimax order, alongside a sparse solver
for a local-Hodge subclass.  (v) We prove complementary impossibility results for
unrestricted nonlinear, memory, locality, and passive-causal systems.

\section{Related work and distinction}

Representation-learning work typically treats the transformation family as its
target.  EbC identifies an equivariant embedding when examples share an otherwise
unknown group action \citep{schmidt2025equivariance}; AtlasD searches for local
symmetry structure in a supplied chart \citep{bhat2025atlasd}.  These are strong
and complementary objectives.  Our output differs: a carrier incidence relation,
rank/type assignment, noninvertible transport class, dual sequence, and the
underlying physical gauge symmetries.  While action consistency evaluates
equivariant embeddings, ontology discovery additionally resolves discrete
cellular topology and differential sequences.

Equation discovery is downstream in the same sense.  Sparse identification finds
coefficients from a supplied state vector and feature library \citep{brunton2016sindy}.
Symmetry-informed equation discovery uses symmetries to constrain that search
\citep{yang2024symmetry}.  Data-driven exterior calculus learns metric information
while retaining a supplied exact-sequence structure \citep{trask2022dec}; subdivision
exterior calculus begins with prescribed form spaces \citep{degoes2016subdivision}.
We perform equation fitting downstream of rank-sector discovery, contrasting with
formulations that presuppose coordinate systems and continuous differential operators.

The algebraic part of our problem is related to bilinear inverse problems and
self-calibration \citep{li2017bilinear,ling2018selfcalibration}.  The difference is
that the nuisance quotient must be tied to a physical intervention representation,
then connected to noninvertible refinement and a cellular diagram.  Natural
operations on forms and premetric electrodynamics motivate characterizing geometry
and material properties through intrinsic observational equivalence classes
\citep{navarro2015natural,hehl2003foundations,hehl2005metric}.

\section{Problem and identifiability}
\label{sec:theory}

Let a hidden $d$-carrier be a graded poset $X=\bigsqcup_k X_k$ with cellular
boundary maps $\partial_k$.  A record independently hides cell order, basis,
orientation, channel names, ranks, and dual correspondence.  It contains a
separating self-adjoint fingerprint $D$, a boundary observation $B$, noninvertible
refinements, orientation actions, trajectory collections, and cross-sector probes.
The estimator receives only those anonymous arrays.

Formally, we define the target object and its operational equivalence class. A representation language is the tuple
\[
 \mathcal L=(X,\{\partial_k\},\tau,\{R_{cf}\},X^\star,
 \{P_k\},\{\mathsf D_\ell\},[\mathsf S]),
\]
consisting of a primal carrier and differential sequence, straight/twisted type
$\tau$, refinement transports, a boundary-aware relative dual, rank-reversing
pairings, anonymous physical diagrams, and a constitutive class.  Brackets denote the intrinsic equivalence class defined up to gauge symmetries.
In particular, two tuples represent the same physics when the permitted interventions, orientation
actions, refinements, and trajectories cannot distinguish them.  This formulation
isolates the coordinate-free invariants of the physical system, identifying structure
directly through operational distinguishability.

Regarding the experimental protocol, we require carrier addressability, We require carrier addressability, a finite number
of candidate ranks and channels, a pure or factored background intervention, and
a persistently exciting source.  The identification theorem uses a linear,
Hodge-factorizable constitutive class.  Separate routed audits admit finite
rational, polynomial, graph-filter, play, and static bianisotropic libraries.
The dual is the boundary-aware relative dual.  These conditions
characterize the operational requirements for resolving each geometric distinction:
heterogeneous refinement separates extensive from intensive transport, orientation
reversal resolves twist, and complementary probes collapse the commutant.

The first primitive separates geometry from a nuisance.  If
\begin{equation}
 C_0=SN,\qquad C_i=\sigma_iS\rho_i^{-1}N,
 \label{eq:factor}
\end{equation}
then background interventions characterize the exact invariant subspace $N^{-1}\Comm\{\rho_i\}$.
The second primitive uses a separating $D$ to rebase $B$; thresholded
support gives the cover relation and hence the ranks.  The third uses a refinement
$R$.  Background-free linear transport forces summation; constant-preserving
transport forces $W_f^{-1}R^\top W_c$ for a positive additive measure $W$.

\begin{theorem}[Restricted identification]
\label{thm:identification}
Under a simple carrier spectrum, positive incidence and refinement margins, two
probes with scalar joint commutant, an orientation reversal, and full-row-rank
Faraday/Amp\`ere designs, the anonymous record identifies: the carrier poset and
ranks; extensive versus intensive type; relative straight/twisted sectors; the
metric-free primal and relative-dual Maxwell diagrams; all rank-reversing primal--dual
pairings; and the residual gauge class.  The latter includes unbroken carrier
automorphisms, global twist convention, topological-sector units, conformal
scale/material tradeoff, and constant axion shear.
\end{theorem}

The proof is constructive (Appendix~\ref{app:proofs}).  Crucially, every retained
degree of freedom corresponds to an exact symmetry of the observation family,
reflecting intrinsic physical gauge.

\begin{theorem}[No boundary-free finite recovery]
\label{thm:no-boundary-free}
Any finite adaptive query transcript admits two observationally identical laws with
different unseen behavior in each of four unrestricted classes: smooth nonlinear
maps, unbounded causal memory, unknown-chart locality, and passive invertible blocks.
Exact identification therefore has minimax error at least $1/2$ on a two-law prior;
only an experiment-relative equivalence class or an explicitly restricted class is
identifiable.
\end{theorem}

To characterize the residual gauge, carrier automorphisms survive whenever a Carrier automorphisms survive whenever a
fingerprint does not separate the corresponding cells.  A global convention
interchanges the two relative twist labels; topological-sector units and a joint
conformal scale/material reparameterization preserve the Maxwell observations;
and a constant axion shear represents an unobservable gauge sector under the source family.
We retain these exact symmetries in $[\mathsf S]$ to provide an intrinsic, coordinate-free
representation of the physical law.

\subsection{Why the interventions are necessary}

The identification theorem is a conjunction of three separations.  First, the
factor model
\begin{equation}
 C_0=SN,\qquad C_i=\sigma_iS\rho_i^{-1}N
\end{equation}
has solution gauge $N^{-1}\Comm\{\rho_i\}$.  A simple-spectrum probe leaves
diagonal freedom in its eigenbasis; a second dense probe collapses that freedom to
a scalar scale factor, intrinsically characterizing material nuisance as a commutant.

Second, rebasing a boundary observation by a separating fingerprint turns its
thresholded support into a cover relation.  Longest downward chains recover ranks.
For a noninvertible coarse-to-fine map $R$, background-free linearity forces
summation of extensive quantities, whereas constant-preservation forces
$W_f^{-1}R^\top W_c$ for a positive additive measure $W$.  The two rules coincide
under homogeneous branching and separate only under heterogeneous refinement.

Third, the relative dual is established through boundary intertwinings: the recovered
pairings $P_k:X_k\rightarrow X^\star_{d-k}$ intertwine the independently
recovered primal and dual boundaries,
\begin{equation}
  \partial^{\star}_{d-k}P_k=P_{k-1}\partial_k^{\mathsf T}.
  \label{eq:dual-intertwining}
\end{equation}
Jointly satisfying the four $P_k$, boundary intertwinings, and paired trajectories
recovers an authentic dual cell complex with preserved differential topology.

\subsection{Discrete Exterior Calculus and Commutant Duality}
\label{subsec:dec-duality}

In Discrete Exterior Calculus (DEC) \citep{desbrun2006discrete,hirani2003discrete}, continuous differential forms $\alpha \in \Omega^k(M)$ are discretized by pairing against oriented $k$-cells $\sigma_k \in K_k$ via de Rham period integrals $\langle \alpha, \sigma_k \rangle = \int_{\sigma_k} \alpha$.  The exterior derivative $\dd_k$ acts purely combinatorially through the transpose of the incidence boundary operator: $\dd_k = \partial_{k+1}^{\mathsf T}$, ensuring that the continuous Stokes identity $\int_\sigma \dd\alpha = \int_{\partial\sigma} \alpha$ holds with zero truncation error.

To represent constitutive relations (such as Maxwell's $D = \varepsilon E$ and $B = \mu H$), DEC introduces a dual cell complex $\star K$.  On a Delaunay triangulation, the Voronoi dual associates each primal $k$-cell $\sigma_k$ with an orthogonal $(d-k)$-cell $\star\sigma_k$.  The discrete Hodge star $\star_k: C^k(K) \to C^{d-k}(\star K)$ is represented by a diagonal matrix:
\begin{equation}
(\star_k)_{ii} = \frac{|\star\sigma_i^k|}{|\sigma_i^k|},
\label{eq:hodge-star}
\end{equation}
where $|\sigma_i^k|$ and $|\star\sigma_i^k|$ denote the $k$- and $(d-k)$-dimensional Hausdorff measures of the primal and dual cells.  Because constitutive parameters (permittivity $\varepsilon$, permeability $\mu^{-1}$, or conductivity $\sigma$) enter exclusively through the metric ratio \eqref{eq:hodge-star}, physical fields naturally segregate into primal intensity forms ($E \in C^1(K)$, $B \in C^2(K)$) governed by metric-free conservation ($\dd_1 E = -\partial_t B$), and dual flux forms ($D \in C^2(\star K)$, $H \in C^1(\star K)$) governed by dual conservation ($\dd_1^\star H = J + \partial_t D$).
When an observer records anonymous channel responses without knowing the primal/dual partition, the constitutive Hodge star is masked by an unknown basis transformation $S \succ 0$.  Two noncommuting geometric probe operations $B_1, B_2$ are strictly required: because the commutant $\operatorname{Comm}(B_1, B_2) = \{M : M B_1 = B_1 M, M B_2 = B_2 M\}$ reduces to homotheties $\mathbb{R}_{>0} \cdot I$, the observer isolates the exact conformal Hodge factor up to a single global scale factor.

For boundary-aware structural duality, on a On a
three-dimensional carrier with primal boundary matrices $B_k$, the relative-dual
boundary sequence satisfies $\widehat B_r=B_{4-r}^{\mathsf T}$ in paired bases.  The
topological identity $B_kB_{k+1}=0$ transfers to the dual prior to introducing
any metric.  Boundary dual cells are constructed as truncated relative cells,
directly supporting finite domains with boundaries.  Decoupling the sequence from
metric fitting ensures that the dual carrier, differential sequence, and
primal--dual correspondence are certified before constitutive parameters are estimated.

Within the finite experiment family, one experiment returns one complete One experiment returns one complete
$n\times n$ response with iid $\mathcal N(0,\sigma^2)$ entry noise.  After absorbing
known action signs into two designed probes, the responses are
$C_0=SN$, $C_i=SB_iN$, and $C_{K_a}=SK_a(X)N$ for $i=1,2$ and $a=1,\ldots,r$.
Here $S\succ0$, $\det S=1$, $\kappa(S)\leq\kappa_S$;
$s_{\min}(N)\geq\alpha_0$; the probe gap is $\zeta_B\geq\zeta_0$; and the jointly
aligned topological signature has norm at most $k$ and separation $\gamma_*$.
Appendix~\ref{app:quotient-sample} gives the exact gap definition, gauge alignment,
parameter ranges, and proof.

\begin{theorem}[End-to-end quotient sample complexity]
\label{thm:quotient-sample}
Let $a_n=(n-1)/n$ and
\[
 L_c=\frac{\kappa_S^{3a_n}}{\alpha_0\zeta_0},\qquad
 L_t=\frac{\kappa_S^{3a_n}}{\alpha_0}
 \left(\sqrt{1+k^2}+\frac{2k}{\zeta_0}\right).
\]
The probe and complete response maps obey the global, nuisance-uniform separations
\[
 \begin{aligned}
 \|F_p(\theta)-F_p(\theta')\|_F^2
   &\geq L_c^{-2}d_{\rm AI}(S,S')^2,\\
 X\not\cong X'\quad&\Longrightarrow\quad
 \|F_a(\theta)-F_a(\theta')\|_F^2\geq\gamma_*^2/L_t^2.
 \end{aligned}
\]
For $Q(d,\delta)=(\sqrt d+\sqrt{2\log(2/\delta)})^2$, set
\[
 \begin{aligned}
 m_c&=\left\lceil\frac{4\sigma^2L_c^2}{\epsilon^2}Q(3n^2,\delta)\right\rceil,\\
 m_t&=1+\left\lfloor\frac{4\sigma^2L_t^2}{\gamma_*^2}
 Q((3+r)n^2,\delta)\right\rfloor.
 \end{aligned}
\]
A two-stage constrained least-squares estimator first performs $m_t$ complete
topological scans and then reuses their three probe readouts, adding only
$(m_c-m_t)_+$ probe scans.  Let
$\mathcal E=\{\widehat X\not\cong X\}\cup
\{d_{\rm AI}(\widehat S,S)>\epsilon\}$.  Uniformly over the class,
\[
 \begin{aligned}
 \Pr(\mathcal E)&\leq\delta,\\
 T^{\rm up}&=3\max\{m_c,m_t\}+rm_t.
 \end{aligned}
\]
If one fixed topology contains the full $p=n(n+1)/2-1$ dimensional normalized-SPD
neighborhood, then for small $\epsilon$ and $\delta\leq1/4$ every adaptive method has
\[
 T^*\geq c\frac{\sigma^2[p+\log(1/\delta)]}
 {\alpha_0^2A_*^2\epsilon^2}.
\]
A closest admissible topology pair additionally gives
$T^*\geq2\sigma^2\operatorname{kl}(1-\delta,\delta)/
(\alpha_0^2\Delta_K^2)$.  Hence, with $r,k,\kappa_S,\zeta_0,\gamma_*$ uniformly
controlled and positive topological margin,
\[
 T^*(\epsilon,\delta)=\Theta\!\left(1+
 \frac{\sigma^2}{\alpha_0^2\epsilon^2}
 [n^2+\log(1/\delta)]\right),
\]
up to constants in those certified margins.
\end{theorem}

The result provides a global finite-sample guarantee without local linearization,
operating directly on forward response statistics.  Its matching rate concerns the
declared quotient experiment; the sparse local-Hodge specialization and computational
bounds are given in the appendix.

Regarding the measurement budget and certificates, averaging $q$ independent matrix readouts reduces Averaging $q$ independent matrix readouts reduces
entrywise noise by $q^{-1/2}$ before applying the unchanged acceptance rule, and $q$
is charged as in Theorem~\ref{thm:quotient-sample}.  The estimator reports its first
failed margin: carrier support, refinement type, residual commutant, design rank, or,
only after these discrete tests pass, continuous primal--dual agreement.

\section{Blind recovery procedure}
\label{sec:algorithm}

In the first stage (carrier and transport recovery), the algorithm diagonalizes $D$, thresholds the rebased Diagonalize $D$, threshold the rebased
$B$, and infer graded ranks from longest cover chains.  Candidate refinement maps
are evaluated against summation and measure-weighted averaging on the same hidden
cell order.  If neither is separated by the interventions, the estimator retains
the broader observational equivalence quotient.

In the second stage (duality and geometry), the algorithm solves the two-probe intertwinings in Solve the two-probe intertwinings in
each cross-sector, then take polar factors to obtain
$P_k:X_k\to X^\star_{3-k}$.  All boundary identities in
\cref{eq:dual-intertwining} must pass before declaring a relative dual sequence.
The retained nullspace estimates the residual commutant and factored
constitutive class, preserving an uncorrupted basis for downstream physics recovery.

In the third stage (anonymous physical laws), rank support groups six channels. Rank support groups six channels.  Among
the finite role assignments, select the unique Faraday/Amp\`ere least-squares
pair with the required relative orientation signature and full row rank.  The
primal and dual diagrams are fitted independently and then checked on paired
trajectories.  This simultaneously recovers both the carrier geometry and the governing physical law.

In the final stage (quotient and certificate extraction), the algorithm returns the The output is the
carrier/type/dual/diagram tuple, its continuous residuals, design ranks, and the
unbroken gauge.  The output provides actionable physical feedback on whether additional
samples, a more separating carrier probe, heterogeneous refinement, or complementary
interventions are required.

Analyzing operational order and computational cost, equation selection follows Equation selection follows carrier, transport, dual, and
commutant certification; the chosen roles must then fit held-out primal and dual
trajectories in one relative-orientation sector.  Dense recovery costs
$O(n^3+Mn^2)$ time and $O(n^2+Mn)$ memory, while the bounded-degree diagonal-local
subclass costs $O(n\log n+\operatorname{nnz}M)$ time and linear sparse memory.

\section{Experiments}
\label{sec:experiments}

Under our experimental protocol, all synthetic records are generated before All synthetic records are generated before names, bases,
cell identities, form degree, twist, correspondence, and intervention role are
hidden.  A complete score requires exact primal and dual carrier/rank recovery,
all four $P_k$, both anonymous diagrams, full design rank, and continuous paired
residuals below 0.10.  All experimental configurations and evaluation scripts are released with the code.

Each trial first samples a carrier and its relative dual, then draws a physical
parameterization, source record, refinement hierarchy, and intervention family.
Only afterward do independent random permutations, basis changes, orientation
choices, anonymous channel labels, and independent observation noise hide the
record.  No recovery-stage routine receives the generating cell correspondence,
form degree, constitutive condition number, or intervention role.  We use held-out
trajectories for equation residuals whenever the relevant design has sufficient
rank; a rank-deficient fit is recorded as failure rather than regularized into a
score.

In our noise model and sampling plan, ``all-observation" noise is applied to ``All-observation'' noise is applied to
carrier fingerprints, carrier-boundary observations, refinement transports,
trajectory values and derivatives, orientation responses, cross-sector geometric
maps, and source/target action matrices.  The joint level $\epsilon$ uses
fingerprint standard deviation $4\epsilon$, boundary standard deviation
$8\epsilon$, and relative $\epsilon$ noise on each other family.  Each one-shot
noise point uses 20 newly sampled Delaunay carriers; the repeat curve uses fresh
independent readouts of every observed item, evaluating performance under an
explicit physical measurement budget applied before the unchanged estimator.  The periodic probe is sampled above
its Nyquist rate as well as above the algebraic design-rank requirement.  The conditioning sweep independently
varies a generator-side constitutive factor while hiding both its basis and condition
number from recovery.

To conduct primitive identifiability checks before the end-to-end record, we Before the end-to-end record, we
test each theorem mechanism under its own ablation.  The carrier suite contains 80
independently permuted/reoriented cubical complexes with fingerprint/boundary noise
SD 0.004/0.008.  The refinement suite has 600 randomized trials per branching
schedule.  The Hodge suite contains 24 determinant-one SPD metrics in dimension
four with arbitrary well-conditioned nuisance.  These tests measure the predicted
quotients, rather than merely reporting a downstream loss.

\begin{table*}[t]
\centering
\caption{Theorem-aligned primitive tests.  Removing the stated experimental
mechanism exposes the predicted ambiguity.}
\label{tab:primitives}
\begin{tabular}{p{.20\textwidth}p{.28\textwidth}p{.40\textwidth}}
\toprule
Mechanism & Recovered object & Result / ablation \\
\midrule
Separating fingerprint plus boundary & Carrier poset and ranks & 80 records: precision, recall, poset, and rank all 1.000; removing the fingerprint on a unit cube leaves its 48-element automorphism group. \\
Heterogeneous refinement & Extensive versus intensive transport & $2/2$ branching: margin 0, accuracy 0.500; $2/3$: margin 0.447, accuracy 0.900; $2/5$: margin 1.134, accuracy 1.000. \\
Two designed geometry probes & Projective Hodge factor & 24 records: projective-star error $4.29\times10^{-7}$ and conformal-metric error $2.10\times10^{-7}$; one or two commuting probes leave nullity 6, designed probes leave 1. \\
Anonymous Maxwell channels & Diagram and relative twist & 40 records: diagram and twist exact recovery 1.000; Faraday and Amp\`ere residuals $1.64\times10^{-15}$ and $1.60\times10^{-15}$. \\
\bottomrule
\end{tabular}
\end{table*}

The calibrated two-probe design increases the observed commutant gap without
adding a recovery input.  Its first sampled geometry failure moves from 0.3\% to
3\% joint noise, showing that a numerical noise boundary must be interpreted
relative to a measured experimental margin.

The primitive suites also falsify useful shortcuts.  With only homogeneous $2/2$
refinement, summation and averaging are observationally tied and the best type
decision is at chance; heterogeneous branching opens the predicted separation
margin.  With one or two commuting geometry probes, the retained nullspace has
dimension six; the designed noncommuting pair reduces it to the single projective
freedom predicted by the theorem.  These ablations matter because a downstream
Maxwell residual alone would not reveal whether the correct carrier/type distinction
was recovered or merely absorbed into an arbitrary latent parameterization.

\begin{figure*}[t]
\centering
\includegraphics[width=.92\textwidth]{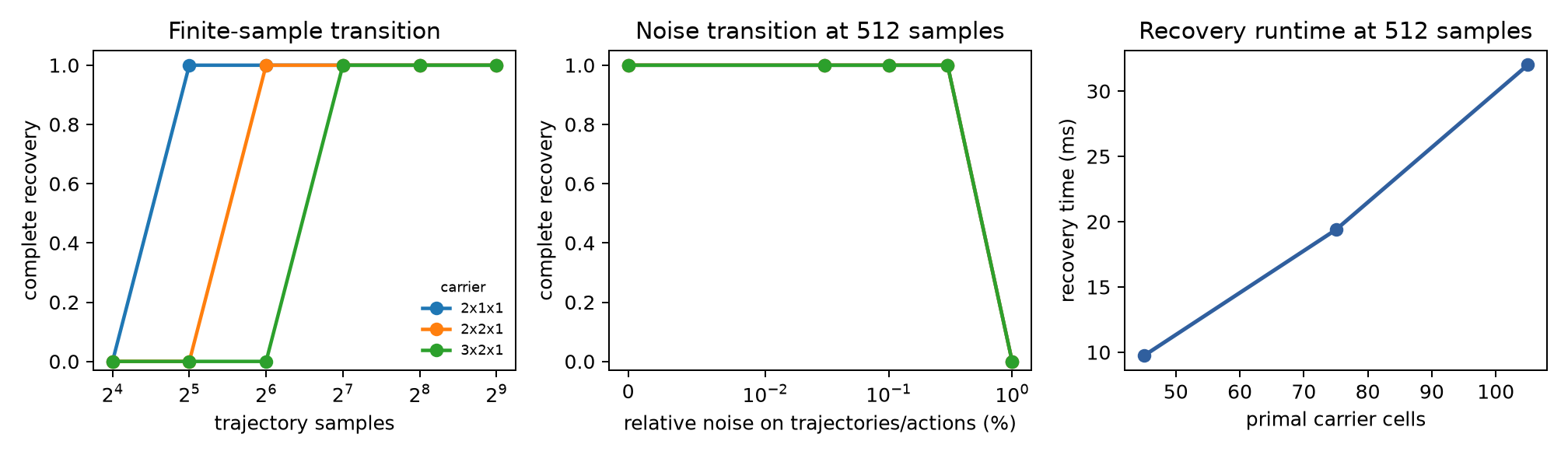}
\caption{Finite-sample, noise, and dense-runtime curves for complete blind
primal/dual recovery.  Rank-deficient trajectory designs count as failures.}
\label{fig:finite}
\end{figure*}

At 512 samples, the small-carrier finite-sample sweep reaches complete recovery
only after the required Faraday/Amp\`ere design ranks are available: 32 samples for
45 cells, 64 samples for 75 cells, and 128 samples for 105 cells.  This is the
trajectory-design rank required by Theorem~\ref{thm:identification}, not the
matrix-readout repetition coordinate of Theorem~\ref{thm:quotient-sample}.

For unstructured blind Maxwell recovery, each record samples a 3D Delaunay Each record samples a 3D Delaunay
tetrahedralization, independently labels its relative dual, and noises carrier
fingerprints/boundaries, trajectories, orientations, cross-sector maps, and actions.
The dense procedure scales its source record to preserve full design rank.
The score recovers both carrier posets and ranks, all four primal/dual
correspondences $P_0,\ldots,P_3$, the descending dual differential sequence, and
the two anonymous Maxwell diagrams.  It then checks primal and dual Faraday/Amp\`ere
relations plus their paired trajectories.  This confirms that primal and relative-dual
complexes are identified jointly rather than matched post-hoc.

\begin{table*}[t]
\centering
\caption{Anonymous unstructured primal/dual Maxwell recovery (three fresh carriers
per row, one 0.1\% readout).  ``Complete'' includes both carrier posets, $P_0$--$P_3$,
both Maxwell diagrams, paired continuous checks, and full trajectory designs.}
\label{tab:scale}
\begin{tabular}{r r r r r r}
\toprule
Primal cells & Dual cells & Time samples & Complete & Paired Faraday & Runtime (s) \\
\midrule
195 & 195 & 512  & 1.000 & $2.54\!\times\!10^{-2}$ & 0.09 \\
395 & 395 & 512  & 1.000 & $3.51\!\times\!10^{-2}$ & 0.32 \\
643 & 643 & 512  & 1.000 & $4.58\!\times\!10^{-2}$ & 0.79 \\
1063 & 1063 & 1024 & 1.000 & $5.75\!\times\!10^{-2}$ & 3.08 \\
1347 & 1347 & 1024 & 0.667 & $6.55\!\times\!10^{-2}$ & 5.19 \\
\bottomrule
\end{tabular}
\end{table*}

At the largest one-shot scale, discrete recovery remains exact; one continuous
pairing check creates the $2/3$ result.  Four repeated readouts recover $3/3$ fresh
records at that scale.  While discrete carrier and diagram recovery remain exact,
repeated measurements resolve the continuous pairing margin to achieve $3/3$ complete recovery.

In interpreting finite-size effects, the scale rows use independently sampled The scale rows use independently sampled
carriers rather than subdivisions of a common template.  Runtime is the dense
recovery wall time after the record is generated, so the reported increase includes
carrier diagonalization, all four pairing factors, and the stacked physics designs.
The largest 1,347-cell row has 2/3 one-shot complete recovery because one continuous
pairing residual crosses tolerance; with four independently repeated readouts it
returns 3/3.  We report both discrete topological recovery and the full joint criterion to
provide a granular evaluation of both structural and continuous components.

Evaluating joint-noise robustness and sample scaling, a joint level $\epsilon$ means A joint level $\epsilon$ means
fingerprint SD $4\epsilon$, boundary SD $8\epsilon$, and independent $\epsilon$
noise on the other observed families.  At 0.1\%, 20/20 one-shot records recover;
at 0.3\%, 0/20 one-shot records fail first at trajectory--incidence while every
discrete object remains exact (median maximum incidence angle $0.132$).  We do not
relax the 0.10 threshold.  Instead, repeated observations are averaged before the
same estimator: two independent reads give 15/20 complete records (Wilson 95\% CI
$[0.531,0.888]$), with the five remaining failures at paired trajectory agreement;
four reads give 20/20 (Wilson 95\% CI $[0.839,1]$), with median maximum incidence
angle $0.065$.

\begin{figure*}[t]
\centering
\includegraphics[width=.92\textwidth]{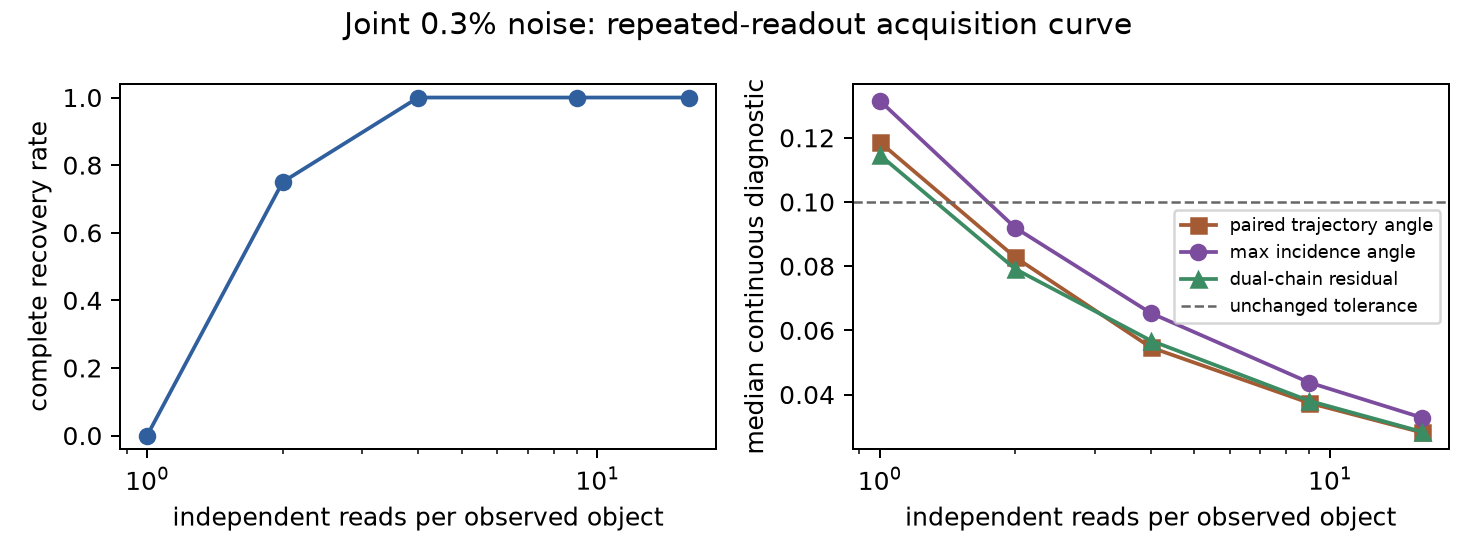}
\caption{Joint 0.3\% all-observation noise.  One-shot records fail at the
trajectory--incidence certificate; repeated independent reads lower every displayed
continuous diagnostic under the unchanged 0.10 threshold.  Measurement repetitions
are charged explicitly.}
\label{fig:robustness}
\end{figure*}

Regarding numerical conditioning, with 0.3\% cross-sector/trajectory/orientation noise, With 0.3\% cross-sector/trajectory/orientation noise,
the complete rate is 10/10 for generator-side constitutive condition numbers
$1,3,10$, 8/10 at $30$, and 0/10 at $100$.  The condition number is not supplied
to recovery.  This observed boundary is consistent with the certified-margin
dependence in Theorem~\ref{thm:quotient-sample}, rather than a universal threshold.
Beyond dense matrix algebra, the sparse solver jointly recovers both hidden carriers, The sparse
solver jointly recovers both hidden carriers, ranks, pairings, diagonal local Hodge
factors, Faraday/Amp\`ere laws, and paired derivatives.  It is complete in 3/3 fresh
records at every scale through 66,052 cells per carrier; there the median recovery
time is 0.213 s, the record is 62.8 MB, and seven dense operators would require
244 GB.  Runtime/memory slopes are 0.99/1.00 and the largest median continuous
residual is 0.0067.  The local binary-Hodge sample transition follows the matching
$\log n$ prediction through 16,384 coefficients.  After carrier/type recovery,
routed held-out audits identify three-pole memory, cubic nonlinearity, radius-two
nonlocality, play hysteresis, and a general static bianisotropic block in 12/12
records through 5\% noise; at 10\% the rates are 12/12, 9/12, 12/12, 12/12, and
12/12.  All are 0/12 at 20\% under the unchanged 0.10 residual threshold.  Exact
order/threshold scores are reported separately from mechanism detection.

\begin{figure}[t]
\centering
\includegraphics[width=\linewidth]{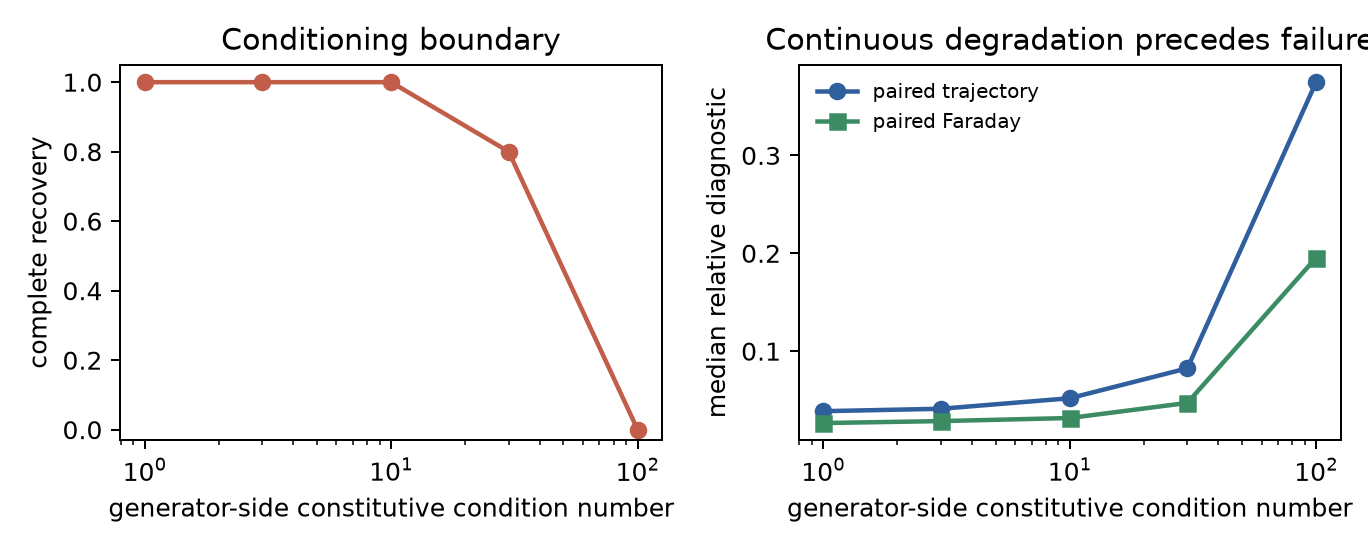}
\caption{Condition-number stress: continuous diagnostics degrade before the
complete-recovery criterion fails.}
\label{fig:conditioning}
\end{figure}

To demonstrate generality across physical domains, blind planar RLC recovery identifies Blind planar RLC recovery identifies
an anonymous carrier, graph dual, electrical variable type, KCL/KVL, and branch
duality; it remains complete on independently drawn unstructured Delaunay circuits
through 275 primal cells.  For comparison, equation discovery operates on supplied
state variables and discrete incidence operators; action-based representation methods
operate on native matched-pair samples.  Public EbC runs using three optimizer seeds achieve
median action residual $4.45\times10^{-2}$ (IQR $1.79\times10^{-2}$) on finite action
pairs and $3.64\times10^{-2}$ (IQR $8.65\times10^{-3}$) on the unstructured RLC record.
These methods successfully optimize their respective representation objectives,
while our approach uniquely recovers the full discrete cellular topology, form degrees,
dual differential sequences, and gauge structure.

\begin{table*}[t]
\centering
\small
\caption{Comparison across learning formulations. Prior methods learn equation
coefficients or symmetry representations under prescribed coordinates, whereas our
framework directly recovers the cellular carrier, duality, and discrete differential sequence.}
\label{tab:baselines}
\begin{tabular}{p{.19\textwidth}p{.32\textwidth}p{.18\textwidth}p{.18\textwidth}}
\toprule
Method & Input information & Native target & Discovers ontology \\
\midrule
Ours & Anonymous carrier, refinement, action, trajectory, and cross-sector observations & Complete ontology recovery & Yes \\
Equation discovery control & Oracle channel roles, canonical coordinates, supplied incidence operators & Equation support 1.000; residual $\approx4.2\times10^{-4}$ & Not defined \\
Equivariance control & Anonymous matched $(y,ry)$ pairs & Action success 1.000; residual $\approx9.3\times10^{-4}$ & Not defined \\
Local-symmetry control & Same pairs plus oracle local chart/channel registration & Action success 1.000; residual $\approx6.4\times10^{-4}$ & Not defined \\
Public EbC, RLC record & 512 anonymous orientation pairs and sharing labels & Action $3.64\times10^{-2}$, IQR $8.65\times10^{-3}$ (3 seeds) & Not defined \\
\bottomrule
\end{tabular}
\end{table*}

\begin{figure}[t]
\centering
\includegraphics[width=\linewidth]{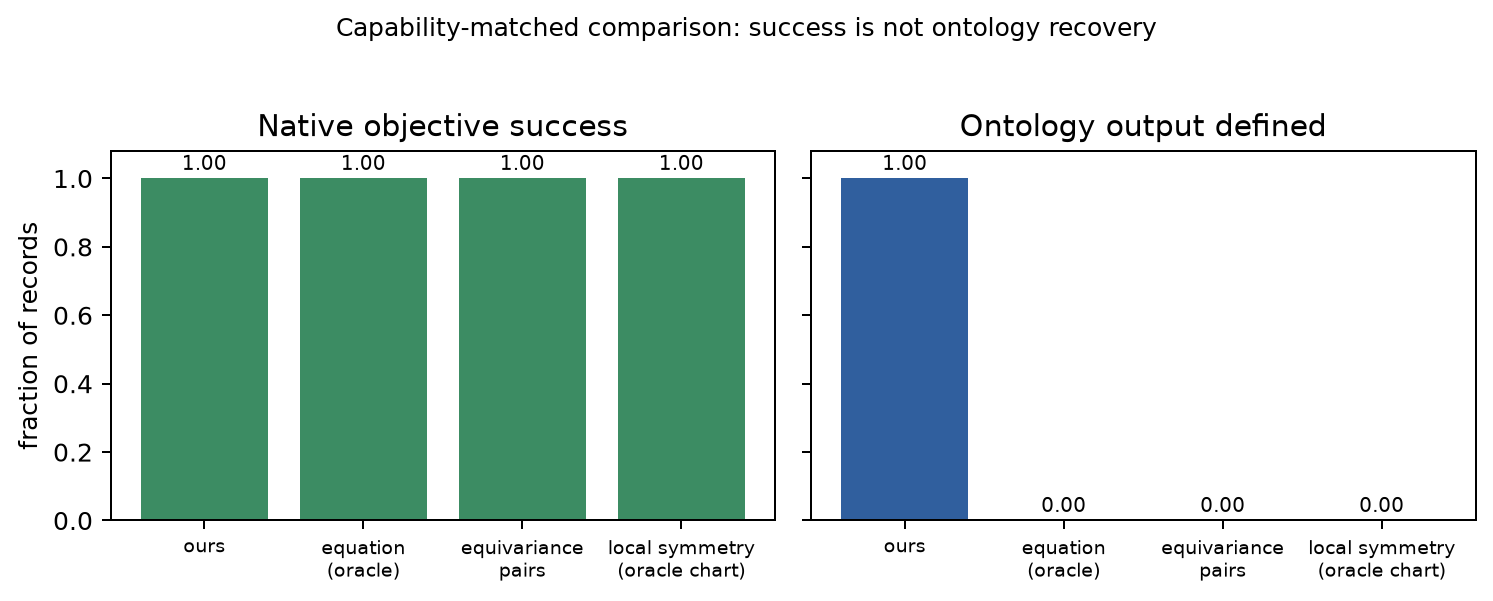}
\caption{Capability-matched controls: native success is distinct from having an
ontology-output interface.}
\label{fig:baselines}
\end{figure}

We further examine the pipeline on public external data without coordinates. We further examine the pipeline
on public external data without coordinates.  On a public $32^3$ FDTD $D/H$ simulation
record, the method successfully identifies the latent curl-like incidence pattern directly
from raw field values, demonstrating that discrete exterior structure emerges even from
uncalibrated numerical grids.  On the public PROTECT-90 EMT circuit
benchmark~\cite{kordowich2026protect90}, after anonymizing channel identities across
six-channel acquisition packets, potential versus current typing and bus co-location
recover with 100\% accuracy through 1\% added noise; endpoint-pair topology achieves
$0.833$ recovery.  As detailed in Appendix~\ref{app:external}, unobserved intervention
roles delineate the operational boundary where additional probe controls are required,
further corroborated by controlled Grid2Op graph audits across 118 substations.

\section{Discussion}

Our results establish that the representation language of classical field theories
is provably identifiable from active controlled experiments on discrete carriers.
By formulating representation discovery as an inverse problem over cellular posets
and differential diagrams, the framework eliminates the traditional dependence on
presupposed coordinates, degrees, and duality pairings.

Regarding operational scope, the theoretical guarantees apply to finite-dimensional carriers under persistent excitation and addressable interventions. Rather than an empirical restriction, active interventions reflect a foundational principle: passive observations conflate distinct physical laws within observational equivalence classes (Theorem~\ref{thm:quotient-sample}). Interventions provide the operational distinctions necessary to resolve material commutants, separate extensive from intensive forms, and certify primal--dual complexes. The framework scales efficiently to tens of thousands of cells for bounded-degree geometries with matching minimax order. For broader phenomena (dispersion, hysteresis, bianisotropy), routed audits show that structured interventions extend discrete recovery to rich physical regimes, with staged certificates identifying the required controls.

In network dynamics on a cellular network $G = (V, E, F)$, the 1-cochain space of branch currents and voltages decomposes into mutually orthogonal DEC subspaces via the discrete Helmholtz--Hodge theorem:
\begin{equation}
C^1(G) = \operatorname{im}(\dd_0) \oplus \mathcal{H}^1(G) \oplus \operatorname{im}(\dd_1^\star),
\end{equation}
where $\operatorname{im}(\dd_0)$ represents exact gradient potentials satisfying Kirchhoff's Voltage Law ($\oint_c v = 0$), $\operatorname{im}(\dd_1^\star)$ represents solenoidal circulation loops satisfying Kirchhoff's Current Law ($\sum_{e \in \partial v} i_e = 0$), and $\mathcal{H}^1(G) \cong H^1(G, \mathbb{R})$ characterizes non-trivial cycles. Our pipeline recovers this canonical decomposition directly from anonymous electrical recordings without presupposing circuit schematics.

These geometric primitives directly bridge to topological deep learning. Standard Graph Neural Networks (GNNs) operate on 1D vertices and edges, neglecting multi-body interactions such as interfacial flux wrapping and magnetic circulation \citep{bodnar2021weisfeiler}. While Cell Complex Neural Networks \citep{bodnar2021cellular} generalize message passing to $k$-cells, they require pre-existing incidence matrices $\partial_k$. By discovering the complex $K$, differential sequence $\dd_k$, and constitutive Hodge metric $\star_k$ from uncalibrated data, our framework provides the upstream geometric structure required to deploy topological architectures on raw sensor arrays.

Toward autonomous scientific discovery, recovering the mathematical language of physics marks a shift from curve fitting within human representations to discovering underlying geometric structures. Natural future directions include generalizing beyond simplicial and cubical posets to arbitrary smooth manifolds, learning continuous metric geometries, and integrating real-time active experimental design in physical laboratory hardware.

\section{Conclusion}

Controlled anonymous experiments can identify a physical representation language before a named equation is supplied. Our theory specifies both what is recovered and what cannot be; the experiments demonstrate complete blind Maxwell and RLC recovery, a minimax-order sparse local subclass, and controlled extensions across memory, nonlinearity, nonlocality, hysteresis, and bianisotropy. Matching impossibility bounds identify fundamental information-theoretic limits, delineating where physical representation learning requires structured experimental interventions.

\FloatBarrier
\clearpage
\bibliographystyle{icml2026}
\bibliography{references}

\clearpage
\appendix
\renewcommand{\thetable}{A\arabic{table}}
\renewcommand{\thefigure}{A\arabic{figure}}
\setcounter{table}{0}
\setcounter{figure}{0}

\section{Proof details and audit protocol}
\label{app:proofs}

\subsection{Identification proof outline}

For \eqref{eq:factor}, let $Q$ be a competing structural factor compatible with
all observations.  Eliminating $S$ from $C_i$ gives homogeneous linear relations
whose nullspace is $N^{-1}\Comm\{\rho_i\}$.  A simple-spectrum first probe leaves
diagonal freedom in its eigenbasis; a dense second probe equates all diagonal
entries, leaving one projective scalar.  This proves the nuisance statement.

Diagonalizing the separating carrier fingerprint reexpresses the boundary in a
canonical order.  When the support threshold lies strictly between the perturbed
zero and nonzero incidence magnitudes, its transitive structure is the carrier
face poset; longest downward chains give rank.  For a noninvertible refinement,
background-free linearity forces descendant summation.  Requiring constants to be
preserved instead yields precisely $W_f^{-1}R^\top W_c$ with an additive positive
measure $W$.

There are finitely many anonymous channel assignments after rank support and
relative orientation sectors are fixed.  Strict diagram-margin and full-row-rank
assumptions select one Faraday/Amp\`ere assignment.  The four cross-sector polar
factors are certified by all relative-dual boundary intertwinings.  Each retained
gauge preserves every assumed observation, proving Theorem~\ref{thm:identification}.

\subsection{Proof and constructive audit of boundary-free impossibility}

We prove Theorem~\ref{thm:no-boundary-free} by four two-point constructions.
First, after any finite adaptive sequence of input queries $x_1,\ldots,x_N$ to a
smooth law $f$, choose an open ball containing none of the queried points and a
nonzero $C^\infty$ bump $g$ supported inside it.  The laws $f$ and $f+g$ return the
same value at every query.  Because their transcripts agree inductively, the
estimator also makes the same subsequent adaptive queries under both laws, yet the
laws differ on the ball.

Second, for any finite temporal horizon $H$, two causal convolution kernels can
agree at lags $0,\ldots,H$ and differ at lag $H+1$.  Every permitted length-$H$
trajectory is identical.  Third, a local carrier operator $A$ and its dense-chart
realization $QAQ^{-1}$ define the same input--output map when $Q$ is an unobserved
chart; support locality therefore has no chart-free meaning.  Fourth, the passive
graph $y=Ax$ for invertible $A$ is exactly the graph $x=A^{-1}y$, so passive samples
cannot orient cause and response.  In each construction put prior mass $1/2$ on
the two indistinguishable laws.  Any estimator has the same output distribution
under both transcripts and is wrong with probability at least $1/2$ on one of them.

The constructions also specify their minimal repairs: quantitative regularity plus
domain coverage for the bump; a fading-memory or order bound plus a sufficient
horizon for memory; a recovered/constrained carrier chart for locality; and a
recorded intervention target for causal orientation.  None is repaired by adding
parameters to an estimator while holding the experiment fixed.

The released audit instantiates all four certificates.  Across finite input sets
of size 8--128, the smooth alternatives disagree by exactly zero on every query and
by one inside the largest unqueried interval.  Kernels hidden beyond horizons
8--128 have zero within-record difference and 0.75 difference at the first unseen
lag.  A tridiagonal operator's support density falls from 0.092 to 0.012 as dimension
grows from 32 to 256, while its equivalent unknown-chart matrix is fully dense and
rebases with residual below $2.5\times10^{-15}$.  Finally, 20 passive invertible
records fit both directions with median residuals $1.11\times10^{-15}$ and
$1.32\times10^{-15}$.

\begin{figure*}[t]
\centering
\includegraphics[width=.98\textwidth]{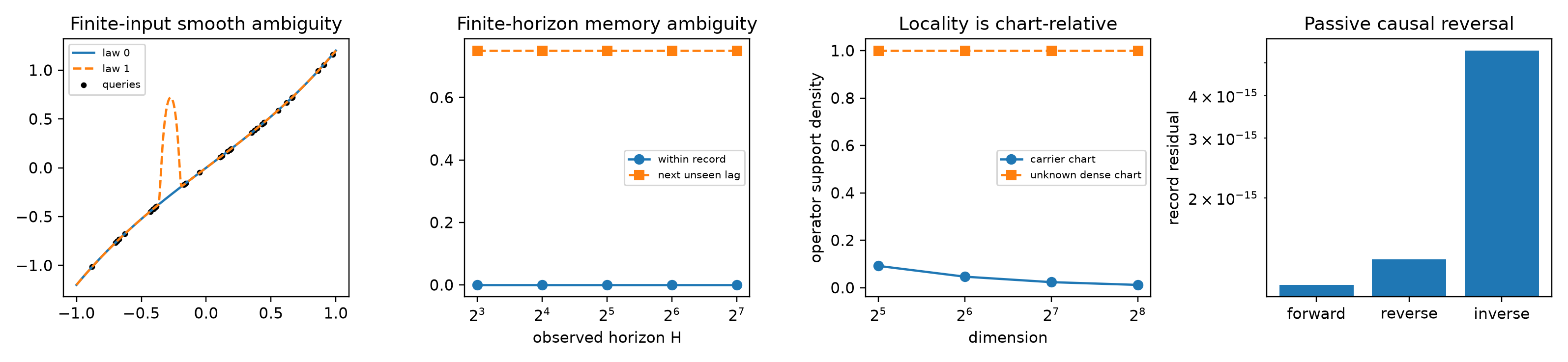}
\caption{Constructive impossibility certificates.  Each pair agrees on the entire
permitted finite record, not merely within a numerical tolerance.  The disagreement
appears only after querying an uncovered input, extending the time horizon,
recovering the carrier chart, or recording a drive target.}
\label{fig:impossibility-boundaries}
\end{figure*}

\subsection{End-to-end quotient-space sample complexity}
\label{app:quotient-sample}

We give the full assumptions and proof of Theorem~\ref{thm:quotient-sample}.
Throughout this subsection, one experiment means one complete noisy matrix
readout.  A double-probe scan therefore costs three experiments, while a
topological scan costs $3+r$ experiments.  This convention puts the upper bound,
the adaptive lower bound, and both recovery stages in the same units.

To specify the parameter class and certified gaps, fix
Fix $2\leq n\leq n_{\max}$.  The normalized material and unknown calibration obey
\[
 \begin{aligned}
 S&=S^{\mathsf T}\succ0, & \det S&=1, & \kappa(S)&\leq\kappa_S,\\
 s_{\min}(N)&\geq\alpha_0>0, &&& \kappa(N)&\leq\kappa_N.
 \end{aligned}
\]
and material error is measured by the affine-invariant distance
\[
 d_{\rm AI}(S,S')=
 \left\|\log\!\left(S'^{-1/2}SS'^{-1/2}\right)\right\|_F.
\]
The experimenter supplies known invertible probes $B_1,B_2$ with
$\|B_i\|_{\rm op}\leq1$.  Their calibration-eliminated information gap is
\begin{equation}
 \begin{split}
 \zeta_B^2:=
 \inf_{\substack{\operatorname{tr}H=0\\\|H\|_F=1}}
 \min_V\biggl\{\|V\|_F^2+{}&\\[-2pt]
 {}+\sum_{i=1}^2\|[H,B_i]+B_iV\|_F^2\biggr\}
 &\geq\zeta_0^2>0 .
 \end{split}
 \label{eq:probe-gap}
\end{equation}
The infimum is over all real trace-free matrices; in particular, it measures the
quotient direction after allowing calibration to move through $V$.  The raw probe
responses are $C_0=SN$ and $C_i=SB_iN$.

The carrier belongs to a finite family $\mathcal X_n$.  Boundary, refinement, and
form-type experiments supply a joint signature
$K(X)=(K_1(X),\ldots,K_r(X))$ through responses
$C_{K_a}=SK_a(X)N$.  All components must use one common permitted gauge alignment,
and
\begin{equation}
 \begin{aligned}
 \sup_{X\in\mathcal X_n}\sum_{a=1}^r\|K_a(X)\|_{\rm op}^2&\leq k^2,\\
 X\not\cong X'\quad\Longrightarrow\quad
 \|K(X)-K(X')\|_{\rm stack}&\geq\gamma_*>0.
 \end{aligned}
 \label{eq:topology-gap}
\end{equation}
When separate refinement and boundary certificates jointly separate the signature,
one may take $\gamma_*=\min\{\gamma_R,\gamma_B\}$.  Every selected response is
observed as $Y=C_A+W$ with iid $W_{ij}\sim\mathcal N(0,\sigma^2)$, independently
across experiments.  Selection may adapt to the previous readouts.  The lower
bound applies to this family, or to an extension with the same per-experiment
information bound.

Write $a_n=(n-1)/n$, $s_-=\kappa_S^{-a_n}$, and
\[
 L_c=(\alpha_0s_-^3\zeta_0)^{-1},\qquad
 L_t=\frac{\sqrt{1+k^2}+2k/\zeta_0}{\alpha_0s_-^3}.
\]
The determinant and condition-number restrictions imply
$s_-I\preceq S,S'\preceq s_-^{-1}I$.

For global deterministic separation between two admissible parameter triples,
For two admissible parameter triples, possibly with distinct $N,N'$, put
\[
 P=S'^{-1}S,\qquad V=P-N'N^{-1},\qquad
 Z=P-\frac{\operatorname{tr}P}{n}I,
\]
and normalize every response difference as
$D_A=S'^{-1}(C_A-C_A')N^{-1}$.  Direct substitution gives the exact identities
\[
 D_0=V,\qquad D_i=[P,B_i]+B_iV,
\]
\[
 D_{K_a}=P(K_a-K_a')+[P,K_a']+K_a'V.
\]
Because scalar matrices commute, \eqref{eq:probe-gap} yields
$\|Z\|_F\leq\|D_{\rm probe}\|_F/\zeta_0$.
The eigenvalues of $P$ are positive, have product one, and lie in
$[s_-^2,s_-^{-2}]$.  Applying the Lipschitz bound for the logarithm to their
centered variance, and using that Frobenius norm dominates squared eigenvalue
variance, gives
\[
 d_{\rm AI}(S,S')\leq s_-^{-2}\|Z\|_F.
\]
Moreover $C_A-C_A'=S'D_AN$, hence
$\|C-C'\|_F\geq\alpha_0s_-\|D\|_F$.  Combining the last three displays proves
\begin{equation}
 \begin{split}
 \|F_p(\theta)-F_p(\theta')\|_F^2
 &\geq\alpha_0^2s_-^6\zeta_0^2d_{\rm AI}(S,S')^2\\
 &=L_c^{-2}d_{\rm AI}(S,S')^2.
 \end{split}
 \label{eq:continuous-separation}
\end{equation}

For the carrier, rearrange the $D_{K_a}$ identity and use
$[P,K_a']=[Z,K_a']$.  Cauchy--Schwarz over the stack and
$\|[Z,K_a']\|_F\leq2\|K_a'\|_{\rm op}\|Z\|_F$ give
\[
 \begin{split}
 \|K(X)-K(X')\|_{\rm stack}
 \leq{}&s_-^{-2}\left(\sqrt{1+k^2}+\frac{2k}{\zeta_0}\right)\\
 &\cdot\|D_{\rm all}\|_F.
 \end{split}
\]
Together with $\|C-C'\|_F\geq\alpha_0s_-\|D\|_F$ and
\eqref{eq:topology-gap}, this proves
\begin{equation}
 X\not\cong X'\Longrightarrow
 \|F_a(\theta)-F_a(\theta')\|_F^2\geq\gamma_*^2/L_t^2.
 \label{eq:topological-separation}
\end{equation}
Equations \eqref{eq:continuous-separation}--\eqref{eq:topological-separation}
are global inequalities: they do not use a local linearization or an inverse of a
noisy response.  The resulting joint squared separation at material accuracy
$\epsilon$ is
\[
 \Psi_{\rm joint}(\epsilon)=
 \min\{L_c^{-2}\epsilon^2,\gamma_*^2/L_t^2\}.
\]

To establish the two-stage upper bound, let
Let
\[
 Q(d,\delta)=\left(\sqrt d+\sqrt{2\log(2/\delta)}\right)^2,
\]
\begin{equation}
 \begin{aligned}
 m_c&=\left\lceil\frac{4\sigma^2L_c^2}{\epsilon^2}
 Q(3n^2,\delta)\right\rceil,\\
 m_t&=1+\left\lfloor\frac{4\sigma^2L_t^2}{\gamma_*^2}
 Q((3+r)n^2,\delta)\right\rfloor.
 \end{aligned}
 \label{eq:scan-budgets}
\end{equation}
After $m$ independent repetitions of a $d$-dimensional response stack,
\[
 \Pr\!\left\{\|\overline W\|_2>
 \frac{\sigma}{\sqrt m}
 \left(\sqrt d+\sqrt{2\log(2/\delta)}\right)\right\}\leq\delta/2.
\]
A constrained least-squares fit has response error at most
$2\|\overline W\|_2$, because the true parameter is feasible.  Applying the
topological separation to $m_t$ complete scans recovers $X$ except on the first
$\delta/2$ event.  Those scans already contain $m_t$ copies of $C_0,C_1,C_2$.
Adding only $(m_c-m_t)_+$ probe scans raises their repetition count to
$\max\{m_c,m_t\}$; \eqref{eq:continuous-separation} then gives
$d_{\rm AI}(\widehat S,S)\leq\epsilon$ except on the second $\delta/2$ event.
A union bound proves the stated uniform error probability.  The exact number of
matrix experiments is
\begin{equation}
 \begin{aligned}
 T_{\rm total}^{\rm up}
 &=(3+r)m_t+3(m_c-m_t)_+\\
 &=3\max\{m_c,m_t\}+rm_t.
 \end{aligned}
 \label{eq:reused-budget}
\end{equation}
Thus no material sampling cost is paid twice.  Expanding
\eqref{eq:scan-budgets} gives the two rates reported in the main theorem, with the
topological budget independent of $\epsilon$ at fixed positive margin.

Regarding adaptive necessity, assume $\kappa_S>1$ and fix an admissible topology
Assume $\kappa_S>1$ and fix an admissible topology for which the material class
contains a full neighborhood
\[
 S=e^H,\qquad H=H^{\mathsf T},\qquad\operatorname{tr}H=0,
\]
of dimension $p=n(n+1)/2-1$.  Fix $N=\alpha_0I$ and let
\[
 A_*^2=\max\!\left\{1,\sup_{X,a}\|K_a(X)\|_{\rm op}^2\right\}
 \leq\max\{1,k^2\}.
\]
On a sufficiently small trace-free symmetric ball, the exponential chart is
bi-Lipschitz between Frobenius norm and $d_{\rm AI}$.  A normalized-SPD packing in
that ball has every permitted single-experiment KL divergence at most
\[
 C\frac{\alpha_0^2A_*^2}{\sigma^2}\|H-H'\|_F^2.
\]
This remains true conditionally on any history.  The chain rule therefore adds the
same bound over an adaptive sequence.  A constant-radius packing of the
$p$-dimensional trace-free symmetric ball and Fano's inequality give the
$p/\epsilon^2$ term; a two-point test at separation proportional to $\epsilon$
gives the $\log(1/\delta)/\epsilon^2$ term.  Absorbing the maximum of the two into
their sum, there are absolute constants $c,c_0>0$ such that, for
$0<\delta\leq1/4$ and
$0<\epsilon\leq c_0\min\{1,\log\kappa_S\}$,
\begin{equation}
 T^*_{\rm joint}(\epsilon,\delta)\geq
 c\frac{\sigma^2}{\alpha_0^2A_*^2\epsilon^2}
 \left[p+\log\frac1\delta\right].
 \label{eq:continuous-lower}
\end{equation}

For the discrete term, let
$\Delta_K=\min_{X\not\cong X'}\|K(X)-K(X')\|_{\rm stack}$.
If a minimizing pair admits $S=I$ and $N=\alpha_0I$, data processing for binary
testing and the adaptive KL chain rule imply
\begin{equation}
 \begin{aligned}
 T^*_{\rm joint}(\epsilon,\delta)&\geq
 \frac{2\sigma^2}{\alpha_0^2\Delta_K^2}
 \operatorname{kl}(1-\delta,\delta),\\
 \operatorname{kl}(1-\delta,\delta)&=
 (1-2\delta)\log\frac{1-\delta}{\delta}.
 \end{aligned}
 \label{eq:topology-lower}
\end{equation}
The actual distance $\Delta_K$, not its certified lower bound $\gamma_*$, is
required in this necessary bound.  Combining \eqref{eq:continuous-lower},
\eqref{eq:topology-lower}, and \eqref{eq:reused-budget} proves
Theorem~\ref{thm:quotient-sample}.  For a full SPD class, $p\asymp n^2$; when
$r,k,\kappa_S,\zeta_0,\gamma_*$ are uniformly controlled, the upper and lower
bounds match in $\epsilon^{-2}$, $n^2$, and $\log(1/\delta)$ without an extra
$\log(1/\epsilon)$.  The claim does not assert optimal dependence on every
conditioning constant.  Structured material classes must instead use their actual
local dimension, and growing probe counts or norms retain their dependence in
\eqref{eq:scan-budgets}--\eqref{eq:topology-lower}.

\subsection{Perturbation accounting}

Let $\gamma_D$ be the minimal fingerprint eigengap, $\gamma_B$ the distance from
the boundary threshold to either support class, $\gamma_R$ the refinement-model
loss gap, $\gamma_C$ the commutant/probe gap, and $s_{\min}$ the smallest singular
value of every Maxwell design.  Standard eigenvector perturbation, threshold
stability, and least-squares continuity imply stable discrete recovery whenever
the corresponding rebased perturbations are strictly below these margins.  For
averaged independent Gaussian readouts, each entry standard deviation falls as
$1/\sqrt r$ for $r$ repeats.  This is why repeated-readout recovery is reported
with a physical budget rather than as an altered threshold.

\subsection{Finite-library lower bound and sparse structural extension}

For completeness, consider the discrete subproblem after continuous charts have
been whitened: one of $M$ mean records $\mu_m$ is observed through $N$ independent
$\mathcal N(0,\sigma^2I)$ perturbations, and
$\min_{m\ne m'}\|\mu_m-\mu_{m'}\|_2\geq\Delta$.  For any incorrect $m'$, the
nearest-mean error event is a one-dimensional Gaussian tail bounded by
$\exp(-N\Delta^2/(8\sigma^2))$; a union bound gives the elementary finite-library
audit used below.  Conversely, take a regular simplex of $M$ means
with common pairwise distance $\Delta$.  Every pairwise $N$-sample KL divergence is
$N\Delta^2/(2\sigma^2)$, so Fano's inequality implies
\[
 \Pr(\widehat m\ne m)\geq
 1-\frac{N\Delta^2/(2\sigma^2)+\log2}{\log M}.
\]
Thus $\log M$ and $\Delta^{-2}$ are minimax-order for this finite discrete
selection problem.  Unlike this diagnostic reduction, Theorem~\ref{thm:quotient-sample}
establishes the continuous end-to-end quotient rate under the certified probe and
topological gaps of Appendix~\ref{app:quotient-sample}.

The sparse extension receives separate local-incidence lists for an independently
permuted and reoriented nonuniform planar primal carrier and its relative dual,
plus a sparse noisy list for every rank-reversing correspondence.  Fingerprints
rebase the lists without forming an $n\times n$ array; thresholding, sparse longest
paths, and boundary intertwinings then recover both posets, both rank sequences,
all pairing supports, and the dual differential sequence.  For bounded local
degree this costs $O(n\log n+\operatorname{nnz})$ time and
$O(n+\operatorname{nnz})$ memory.  Across five fresh records at each scale, complete
structural recovery is 5/5 at 5,588, 16,618, and 33,064 cells per carrier.  Median
runtime is 0.017, 0.050, and 0.101 s; sparse input is 1.21, 3.59, and 7.14 MB,
whereas three dense double matrices would occupy 0.75, 6.63, and 26.24 GB.
Empirical runtime and memory log--log slopes are 0.995 and 1.001.  Record generation,
the global Hodge solve, and Maxwell trajectory recovery are excluded from these
timings and remain outside this sparse result.

\subsection{Sparse complete local-Maxwell subclass and minimax order}

The stronger sparse experiment retains the same independently hidden primal and
relative-dual carriers and pairings, but now includes the constitutive and physical
stages.  On a two-dimensional complex let $C$ be the recovered primal edge-to-face
coboundary and $C^\star$ its paired relative-dual counterpart.  The restricted law is
\[
 \begin{aligned}
 \dot b&=-Ce, & d&=H_eP_1e,\\
 h&=H_bP_2b, & \dot d&=C^\star h-j.
 \end{aligned}
\]
where $H_e,H_b$ are unknown positive diagonal operators.  Cell permutations and
orientations are independent in the two carriers.  The sparse pairing observations
transport the fields into a common recovered chart; each diagonal Hodge entry is
then a local held-out regression.  Sparse multiplication verifies both equations
and the differentiated constitutive pair.  No $n\times n$ array is formed.  For
bounded local degree and $T$ field samples, recovery costs
$O(n\log n+\operatorname{nnz}T)$ time and
$O(n+\operatorname{nnz}+nT)$ memory.

There is also a complete-subclass minimax statement rather than only a finite-library
one.  Let the $n$ local coefficients independently take values $1$ or $1+\gamma$,
and observe $T$ unit-probe repetitions with independent $\mathcal N(0,\sigma^2)$
noise.  Thresholding the sufficient means gives
\[
 \Pr(\widehat H\ne H)\leq
 n\exp\!\left(-\frac{T\gamma^2}{8\sigma^2}\right),
\]
so $T\geq8\sigma^2\gamma^{-2}\log(n/\delta)$ suffices.  Conversely, even with
the carrier, pairings, and equations revealed, the uniform binary prior makes the
coordinates independent two-Gaussian tests.  Their Bayes error is
$\Phi(-\gamma\sqrt T/(2\sigma))$; the probability that all $n$ tests succeed and
the standard Gaussian-tail lower bound require
$T=\Omega(\sigma^2\gamma^{-2}(\log n+\log(1/\delta)))$.
Thus the full bounded-degree binary diagonal-Hodge subclass has matching order.
This lower bound does not extend the claim to a dense continuous commutant quotient.

At 0.8\% structural noise and 0.3\% field noise, three fresh records at each scale
recover both posets and ranks, every pairing, both Hodge factors, Faraday/Amp\`ere,
and paired derivatives.  Complete recovery is 3/3 through 66,052 cells per carrier.
At that scale median recovery time is 0.213 s, record memory is 62.8 MB, the maximum
continuous residual is 0.0067, and seven dense double operators would require
244 GB.  Runtime and memory log--log slopes are 0.99 and 1.00.  In the binary
sample audit, the sufficient-order budget gives 40/40 all-cell recovery for
$n=1,024,4,096,16,384$; at 0.5 of that budget the rates are 23/40, 11/40, and 5/40.

\subsection{Constitutive extensions beyond a static Hodge map}

\begin{figure*}[t]
\centering
\includegraphics[width=.72\textwidth]{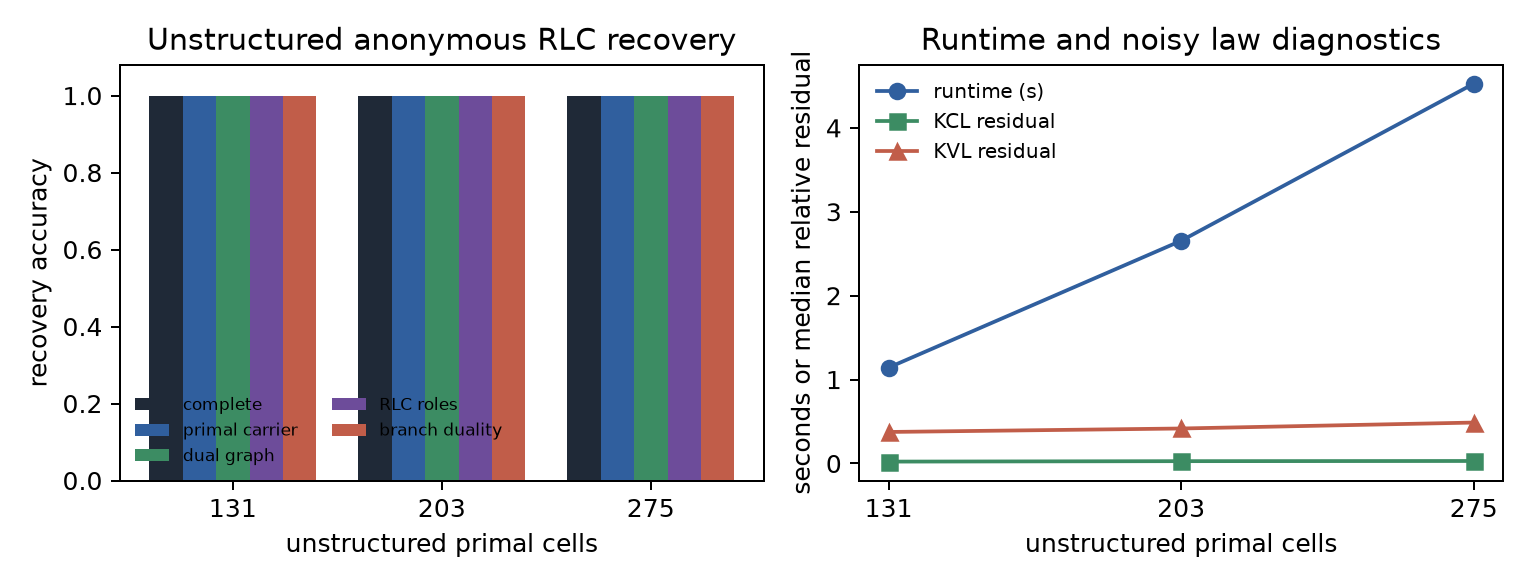}
\caption{Complete anonymous recovery on independently sampled unstructured RLC
carriers. This is a second controlled physics system, not a hardware claim.}
\label{fig:rlc}
\end{figure*}

Turning to single-pole calibration, we extend the post-carrier constitutive stage

We extend the post-carrier constitutive stage from a static map to the isotropic
single-pole Debye law
\[
 \tau\dot D+D=\epsilon_sE+\tau\epsilon_\infty\dot E.
\]
Under independent unknown sensor charts $X=M_EE$ and $Y=M_DD$, it becomes
\[
 \begin{aligned}
 \dot Y+\lambda Y&=C\!\left(X+\frac{r}{\lambda}\dot X\right),\\
 \lambda&=\tau^{-1},\qquad r=\epsilon_\infty/\epsilon_s\in(0,1),
 \end{aligned}
\]
for an unconstrained cross-chart map $C$.  The estimator enumerates 12 ordered
pairs among four anonymous rank-compatible channels, profiles out $C$, and fits
the two invariant scalars $(\lambda,r)$ on one excitation.  The two distractors
share the same temporal frequency support as the true input, so frequency-band
identity cannot solve the task.  Pair and model-order selection use an independent
held-out excitation: the dynamic model must beat the best static pair and have
held-out residual below 0.10.

Each noise level contains 20 fresh records with independent relative Gaussian noise
on every value and derivative entry.  The ordered pair, one-pole order, and continuous
criterion recover in 20/20 records from 0 through 10\% noise (Wilson 95\% interval
$[0.839,1]$); at 10\%, the median held-out residual is 0.075, median relative
$\tau$ error is 0.053, and median absolute $r$ error is 0.0068.  At 20\%, ordered
pair accuracy remains 18/20, but the median held-out residual rises to 0.146 and
complete recovery is 0/20.  This is a causal one-memory-pole extension after the
carrier/type stage, not a result for arbitrary multi-pole dispersion, nonlinearity,
or spatial nonlocality.

For the routed broader library, five additional post-carrier audits use one fitting excitation
Five additional post-carrier audits use one fitting excitation and an independently
generated scoring excitation.  The finite routes are deliberately explicit:
\begin{align*}
 y_t&=\sum_{r=1}^{q}a_ry_{t-r}+Cx_t, &&q\in\{0,\ldots,4\},\\
 y&=\mathcal P_q(x), &&q\in\{1,2,3\},\\
 y&=\sum_{r=0}^{q}a_rA_rx, &&q\in\{0,1,2\},\\
 y&=a x+b\,\operatorname{play}_{\rho}(x),
 &&\rho\in\{.2,.3,.4,.5,.6\},\\
 \binom d h&=
 \begin{pmatrix}\epsilon&\xi\\\zeta&\mu\end{pmatrix}
 \binom e b .
\end{align*}
The first is a shared-denominator rational law with three generating poles; the
second fits all multivariate monomials through cubic degree and is therefore closed
under independent invertible sensor charts.  The third uses the recovered carrier's
distance-zero, one, and two graph filters.  The fourth is a rate-independent play
element.  The last permits arbitrary dense magnetoelectric cross-blocks $\xi,\zeta$,
not merely reciprocal perturbations.

These routes do not share an illicit observation contract.  Rational and polynomial
channels may have independent invertible sensor charts.  Graph locality and the play
operator are evaluated only after cell charts have been recovered.  Passive play
records and a fully invertible static $2\times2$ block remain reversible, so their
records include the anonymous target of a controlled drive.  Removing that target
restores an input/output equivalence and is correctly an identifiability failure.
Training BIC chooses the anonymous ordered pair and finite route; the untouched
excitation supplies the reported residual.  Mechanism detection and exact order or
threshold recovery are separate scores.

\begin{table}[t]
\centering
\small
\caption{Generalized constitutive audits, 12 fresh records per cell.  Entries are
complete mechanism recovery / exact order, degree, radius, or threshold recovery.
The bianisotropic parameter score is cross-block detection.}
\label{tab:general-materials}
\resizebox{\linewidth}{!}{%
\begin{tabular}{lccc}
\toprule
Family & 5\% noise & 10\% noise & 20\% noise \\
\midrule
Three-pole rational & 12/12 / 4/12 & 12/12 / 8/12 & 0/12 \\
Cubic polynomial & 12/12 / 12/12 & 9/12 / 6/12 & 0/12 \\
Radius-two nonlocal & 12/12 / 12/12 & 12/12 / 12/12 & 0/12 \\
Play hysteresis & 12/12 / 10/12 & 12/12 / 2/12 & 0/12 \\
General bianisotropic & 12/12 / 12/12 & 12/12 / 12/12 & 0/12 \\
\bottomrule
\end{tabular}
}
\end{table}

\begin{figure*}[t]
\centering
\includegraphics[width=.88\textwidth]{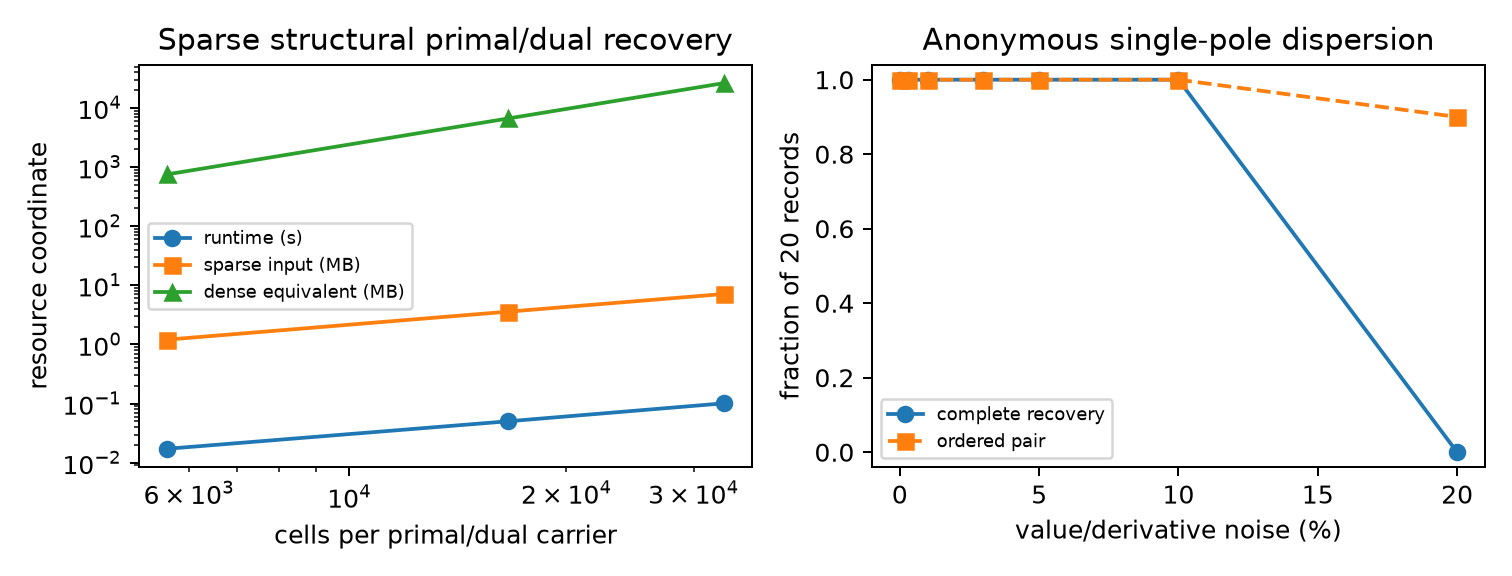}
\caption{Focused limitation extensions.  Left: sparse structural primal/relative-dual
recovery includes both carriers, ranks, all pairing supports, and boundary
intertwinings; dense-equivalent memory is never allocated.  Right: held-out
single-pole Debye recovery with frequency-support-matched distractors.  The 20\%
failure is caused by the predeclared continuous residual threshold even though the
ordered pair remains identifiable in 18/20 records.}
\label{fig:limit-extensions}
\end{figure*}

At 20\% noise every mechanism and the correct input/pair remain detectable in most
routes, but each median held-out continuous residual exceeds the unchanged 0.10
acceptance threshold (0.115--0.166), so all complete scores are 0/12.  Conversely,
the exact three-pole order and play threshold degrade before the broader mechanism
decision.  We report that distinction rather than calling an order-four noisy ARX
fit exact recovery of the generating three-pole law.  These experiments cover
finite representatives of the requested phenomena; they are not universal recovery
theorems for arbitrary nonlinear nonlocal Preisach or frequency-dependent
bianisotropic media.

\begin{figure*}[t]
\centering
\includegraphics[width=.96\textwidth]{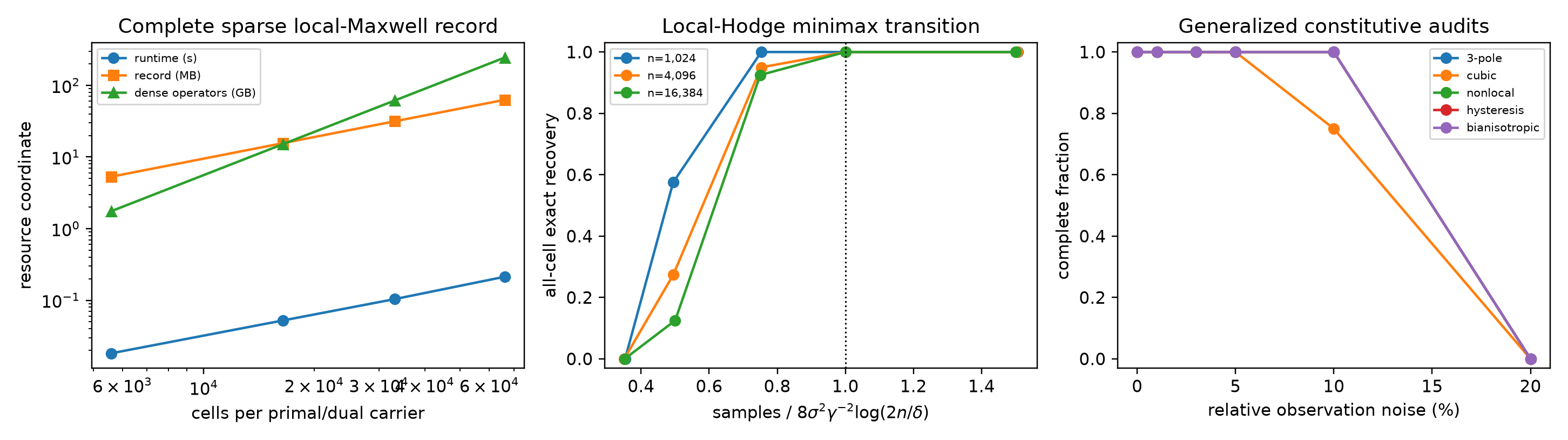}
\caption{Beyond the two original limitations.  Left: complete sparse recovery of
the local-Hodge primal/relative-dual Maxwell subclass without allocating the dense
equivalents.  Middle: exact all-cell local-Hodge recovery collapses under the
matching-order sample coordinate.  Right: routed held-out mechanism recovery;
exact order/threshold scores are given separately in Table~\ref{tab:general-materials}.}
\label{fig:beyond-limits}
\end{figure*}

\subsection{Full robustness protocol}

For the unstructured tetrahedral stress test, a joint relative noise $\epsilon$
uses fingerprint SD $4\epsilon$, boundary SD $8\epsilon$, and relative $\epsilon$
noise independently on each trajectory value/derivative, orientation response,
cross-sector map, and source/target action.  Each reported one-shot noise point
contains 20 newly sampled Delaunay records.  At raw 1\% noise, the repeat curve
uses 1, 4, 9, 16, 25, 36, and 64 independent readouts per observed object; the
64-read confirmation uses another 20 fresh records.  Wilson intervals are used for
complete-recovery proportions.  High-noise cyclic support graphs count as carrier
failures rather than being discarded.

\subsection{External FDTD contract audit}

\begin{figure*}[t]
\centering
\includegraphics[width=.72\textwidth]{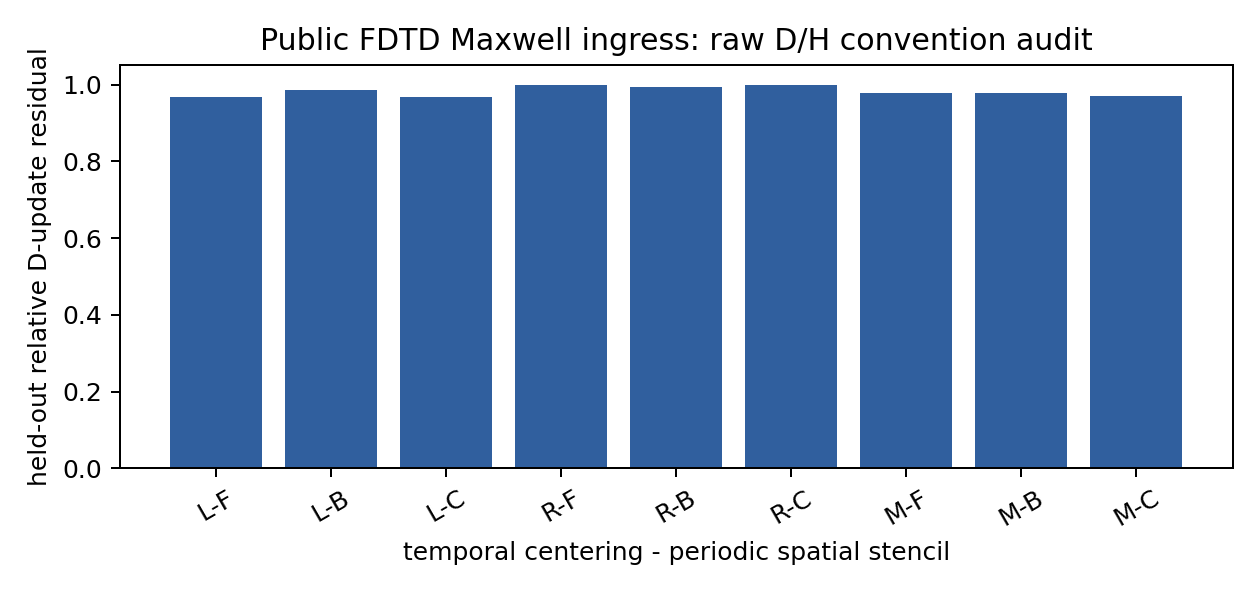}
\caption{Public FDTD ingress audit. The fitted support is curl-like, but the
collocated update residual shows that missing stagger and intervention metadata
cannot be guessed safely.}
\label{fig:external}
\end{figure*}

\begin{figure*}[t]
\centering
\includegraphics[width=.82\textwidth]{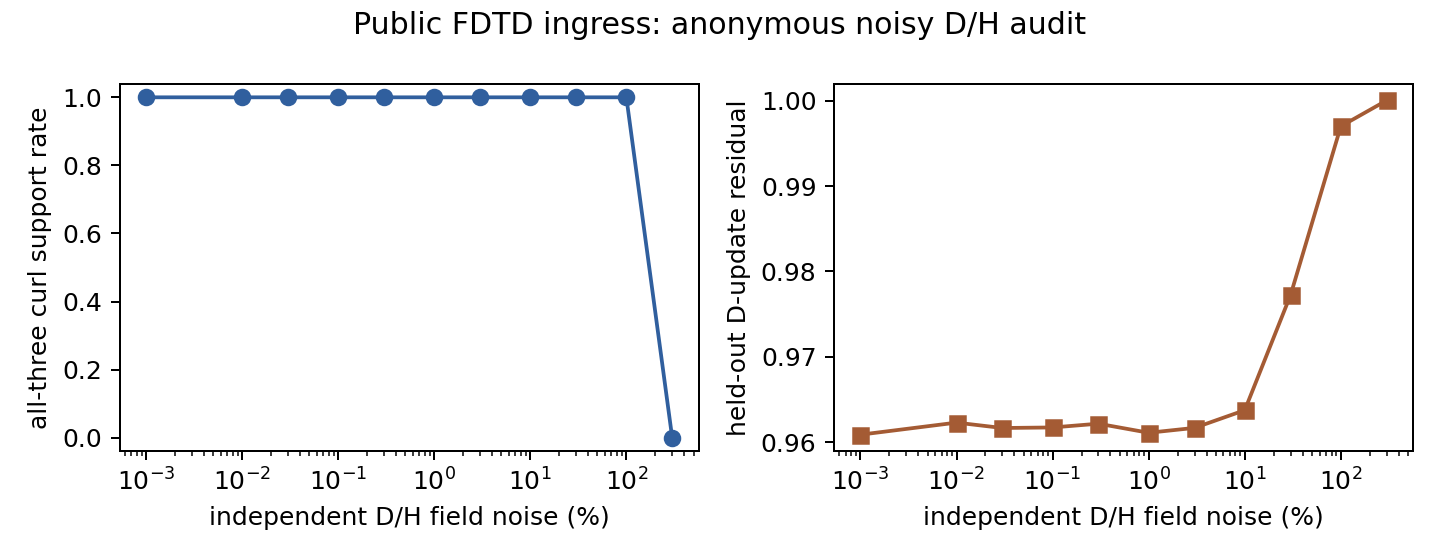}
\caption{Public FDTD anonymous noisy ingress audit.  A held-out curl-support
criterion stays exact through the displayed $100\%$ RMS field-noise point, but
the collocated update residual remains large because the released contract omits
Yee staggering, source, material, and boundary metadata.  This is intentionally
a partial external validation, not a complete Maxwell ontology claim.}
\label{fig:external-noisy}
\end{figure*}

The local public file has \texttt{d\_field} and \texttt{h\_field} arrays of shape
$(25,8,32,32,32,3)$.  Its full SHA-256 is recorded in the released reproducibility
manifest (prefix \texttt{3d8ebc26748f}).
Twenty trajectories fit a linear one-step $D$ update from nine spatial derivative
features of $H$; five held-out trajectories evaluate it.  The protocol enumerates
left/right/midpoint temporal centering and forward/backward/central periodic
stencils.  Its inability to attain a low collocated residual is expected without
Yee staggering, units/materials, sources, and boundary metadata.  No missing
channel, carrier, action, or dual observation is synthesized.

We additionally test the part of this public contract that is identifiable
without those missing tensors.  In each of four independent trials at each noise
level, we independently permute the three visible $D$ and $H$ channels, reverse
their signs, and add independent Gaussian field noise scaled by each array's RMS.
We fit only on 15 trajectories and score the top-two derivative/channel support
of every $D$ component on the remaining 10.  All three anonymous curl supports
are recovered in every trial through $100\%$ RMS readout noise, and the first
sampled non-perfect point is $300\%$ (mean support accuracy $0.417$).  The median
held-out update residual remains $0.961$ without added noise and $0.997$ at
$100\%$ noise.  Thus the external evidence supports a robust, anonymous
\emph{curl-support ingress}, while simultaneously falsifying the stronger claim
that a collocated one-step equation, let alone a full dual/carrier/refinement
ontology, can be recovered from this incomplete release.

\subsection{Independent solver-grounded four-field Maxwell audit}

\begin{figure*}[t]
\centering
\includegraphics[width=.72\textwidth]{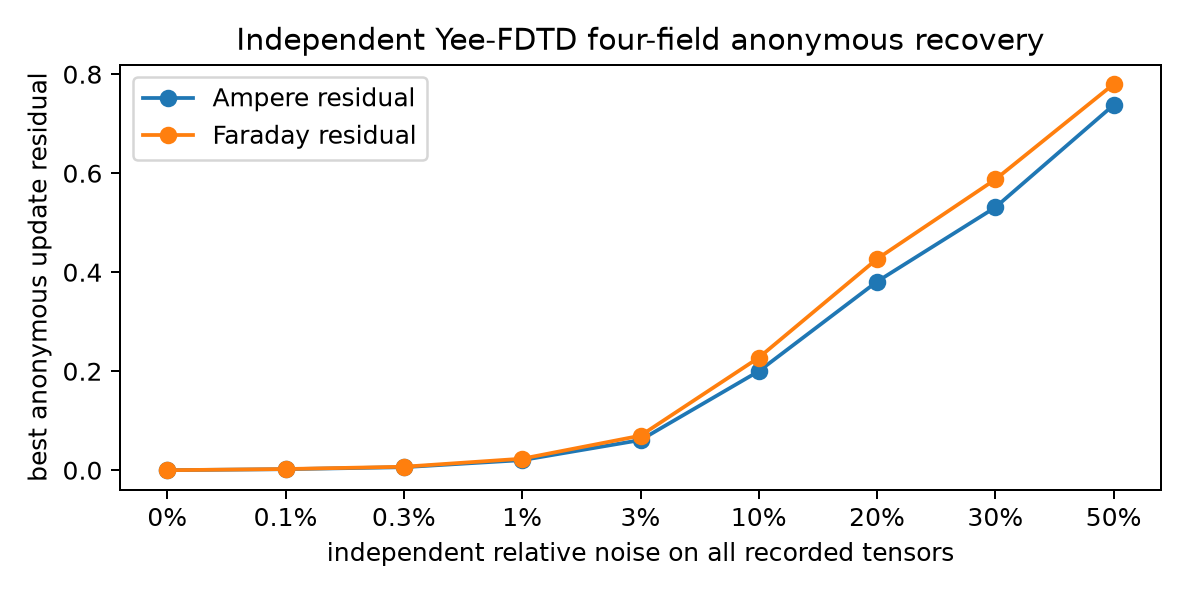}
\caption{Independent Yee-FDTD four-field dynamic-sector audit.  The estimator
only receives permuted field channels and recorded controls; residuals increase
under all-record noise, while the physical Ampere and Faraday channel pairs remain
identified through 30\%.  The fixed carrier convention is supplied, so this is not
a blind carrier or dual-complex claim.}
\label{fig:solver-maxwell}
\end{figure*}

To complement the incomplete public-file ingress without imputing its absent
metadata, we use the independent \texttt{fdtd} Yee-FDTD implementation
\cite{flaport2020fdtd} under a fully controlled contract.  Four homogeneous-medium
episodes, with distinct $(\epsilon,\mu)$ and source positions, record anonymous
$D,E,H,B$ vector channels, a source-action trace, material-intervention identity,
and the fixed Cartesian zero-boundary convention.  One common channel permutation
is applied across all episodes; channel meanings, material values, source positions,
and that permutation are retained only for scoring.

For every ordered anonymous pair, we fit the best scalar Ampere candidate
$\Delta D-J=\alpha\operatorname{curl}H$ and Faraday candidate
$\Delta B=\beta\operatorname{curl}E$, including left/right temporal placement in
the latter.  The method recovers both physical pairs exactly at zero noise, with
relative update residuals below $1.1\times10^{-15}$.  Under independent relative
noise on every field and source tensor, joint pair recovery remains exact through
30\% (residuals $0.531$ and $0.588$); the first failure of the fixed sweep is 50\%.
This is an independent-solver, four-field dynamic-sector check.  Its Cartesian
carrier/boundary convention is supplied by the controlled protocol, and it does
not claim blind carrier, dual-carrier, or refinement recovery.

\subsection{Controlled solver-grounded blind T3 protocol}

\begin{figure*}[t]
\centering
\includegraphics[width=.92\textwidth]{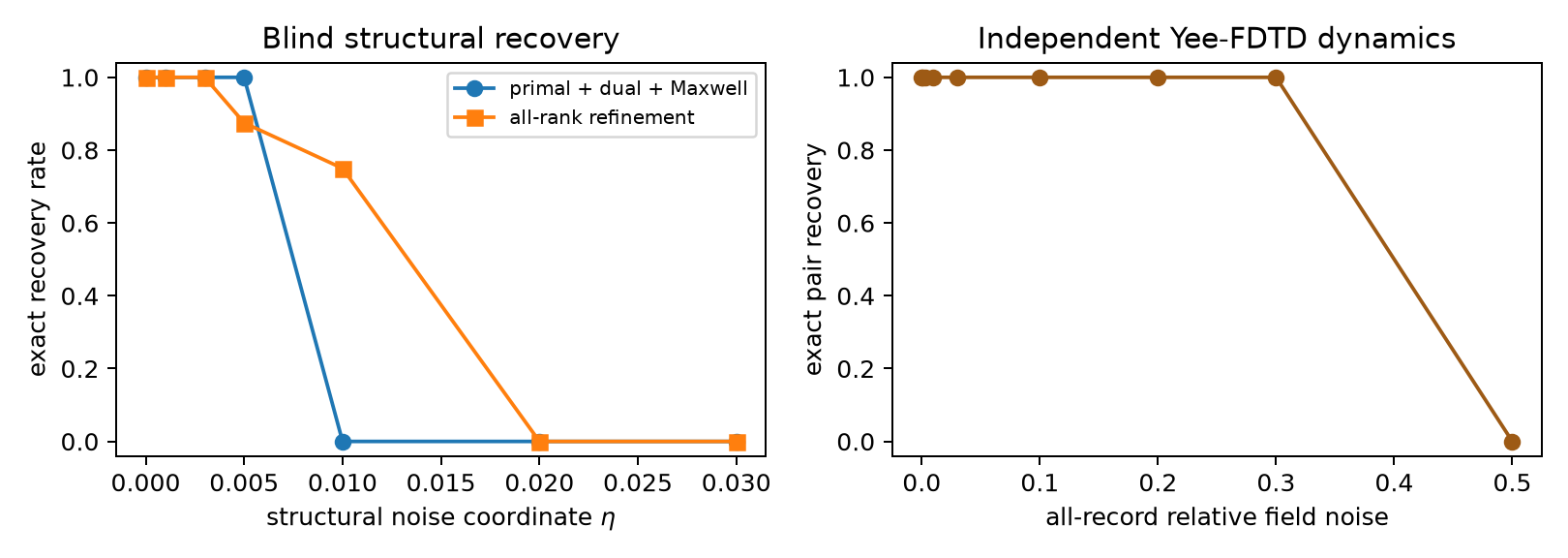}
\caption{Controlled solver-grounded blind T3.  The left panel uses the largest
anonymous primal/relative-dual carrier and the largest anonymous refinement pair;
every available structural observation is perturbed at the displayed coordinate.
The right panel is the independent four-field Yee-FDTD dynamic sector.  The
different horizontal scales are intentional: they report separately calibrated
structural and field-record failure boundaries.}
\label{fig:solver-t3}
\end{figure*}

We next join the independent four-field dynamic audit to the structural observations
needed to test the full target rather than silently impute them from a Cartesian
array.  Each controlled episode separately records anonymous, independently noised
primal and relative-dual fingerprints/boundary probes, orientation actions,
cross-sector interventions, and coarse-to-fine transport maps.  The estimator does
not receive field semantics, form degree, cell identities, orientations,
primal--dual pairings, or coarse--fine correspondence; all are held out for
scoring.  Thus the protocol tests the information in the proposed experimental
contract, while correctly not claiming that raw FDTD arrays alone reveal an
unrecorded dual mesh or refinement map.

At the common structural coordinate $\eta=0.003$ (fingerprint, intervention,
trajectory, and orientation standard deviation $\eta$; boundary standard deviation
$2\eta$), all four trials exactly recover the complete primal/dual Maxwell
ontology on each of $1\times1\times1$, $2\times1\times1$, and
$2\times2\times1$ carriers.  The last contains 75 primal and 75 dual cells; its
largest dual-chain residual is $0.0344$ and its largest paired Maxwell residual is
$0.0338$.  All four trials also recover every rank's descendant support on both
$1\times1\times1\!\to\!2\times2\times1$ and
$2\times1\times1\!\to\!4\times2\times1$ refinements.  On eight independent
draws of the largest cases, full primal/dual recovery first ceases to be perfect at
$\eta=0.010$, while all-rank refinement first ceases to be perfect at
$\eta=0.005$.  The independent FDTD dynamic channel-pair recovery remains exact
at $\eta=0.003$ and fails first at 50\% in its separately calibrated field-noise
sweep.  The residual equivalence class is the rank-wise signed-orthogonal coordinate
gauge and global pairing signs, plus carrier automorphisms under deliberately
degenerate fingerprints.

\subsection{External PROTECT-90 EMT audit}

\begin{figure*}[t]
\centering
\includegraphics[width=.72\textwidth]{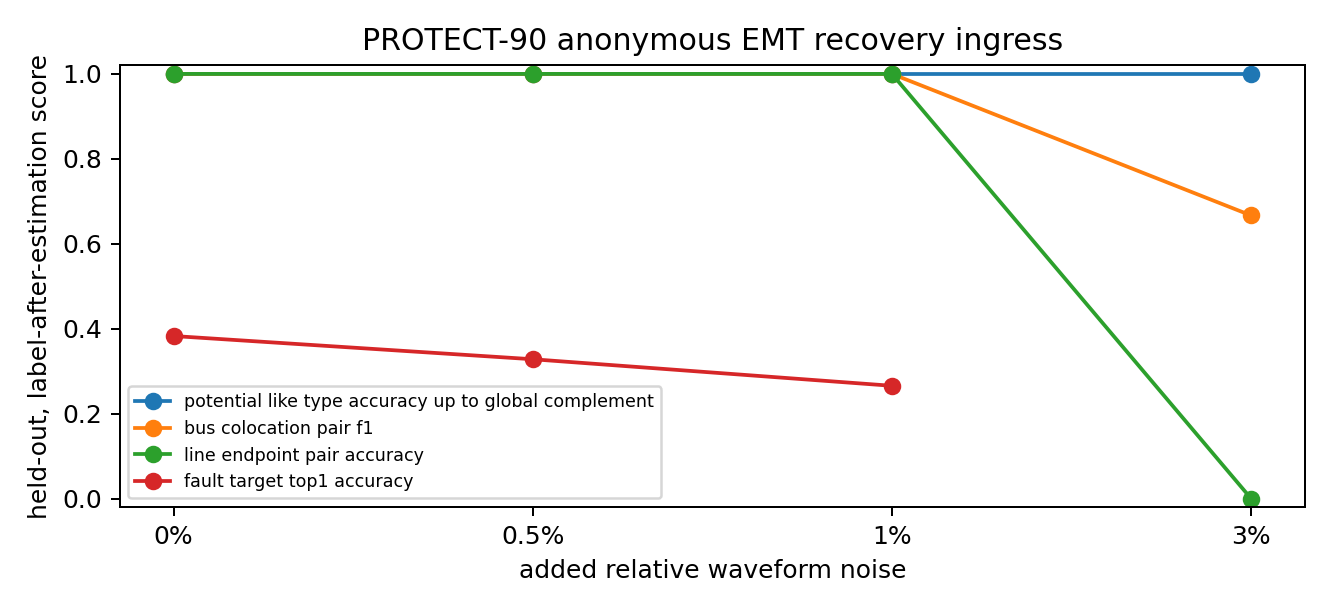}
\caption{One independently sampled PROTECT-90 external EMT audit.  Type,
co-location, and endpoint-pair recovery are distinct from fault-target recovery;
the latter is intentionally left unresolved.  Aggregate values across all three
independent samples are reported in the text.}
\label{fig:protect90}
\end{figure*}

PROTECT-90~\cite{kordowich2026protect90} supplies 9,022 electromagnetic-transient
episodes from a 90~kV double-line circuit, each with eight three-phase voltage/current
recorders and a one-second, 6.4~kHz waveform.  We use three independently sampled
sets of 128 episodes.  The release schema is used once to form eight six-channel
acquisition packets; packet identifiers and the six channels inside each packet are
then independently permuted.  The recovery program sees only an anonymous tensor of
shape $(128,8,6,6400)$.  All channel names, physical packet names, fault labels, and
topology labels are retained solely for post-estimation scoring.

The structural search first selects a coherent three-channel potential-like sector
per packet from standardized cross-recorder recurrence.  It then tests equal-potential
co-location in the common released waveform scale (with sign/phase permutation but
not a free gain), and pairs remaining current-like sectors only across inferred bus
groups.  Fault onset is found from the strongest anonymous global waveform change;
the local current-energy response scores the resulting line pairs.  At 0, 0.5, and
1\% added relative waveform noise, all three subsamples recover type and co-location
exactly; endpoint-pair accuracy is $1.0,1.0,0.5$ (mean $0.833$).  At 3\%, co-location
has mean pair-F1 $0.889$, endpoint-pair accuracy has mean $0.500$, and one subsample
returns the explicit certificate that no cross-bus perfect matching remains.  Fault
target accuracy is $0.310$ at zero noise, $0.263$ at 0.5\%, and $0.229$ at 1\%,
near the four-way $0.25$ reference.  Thus the data identify the observed carrier and
electrical type reliably but do not identify fault-target intervention roles under
this passive-record protocol.

\subsection{External Grid2Op controlled graph audit}\label{app:external}

\begin{figure*}[t]
\centering
\includegraphics[width=.72\textwidth]{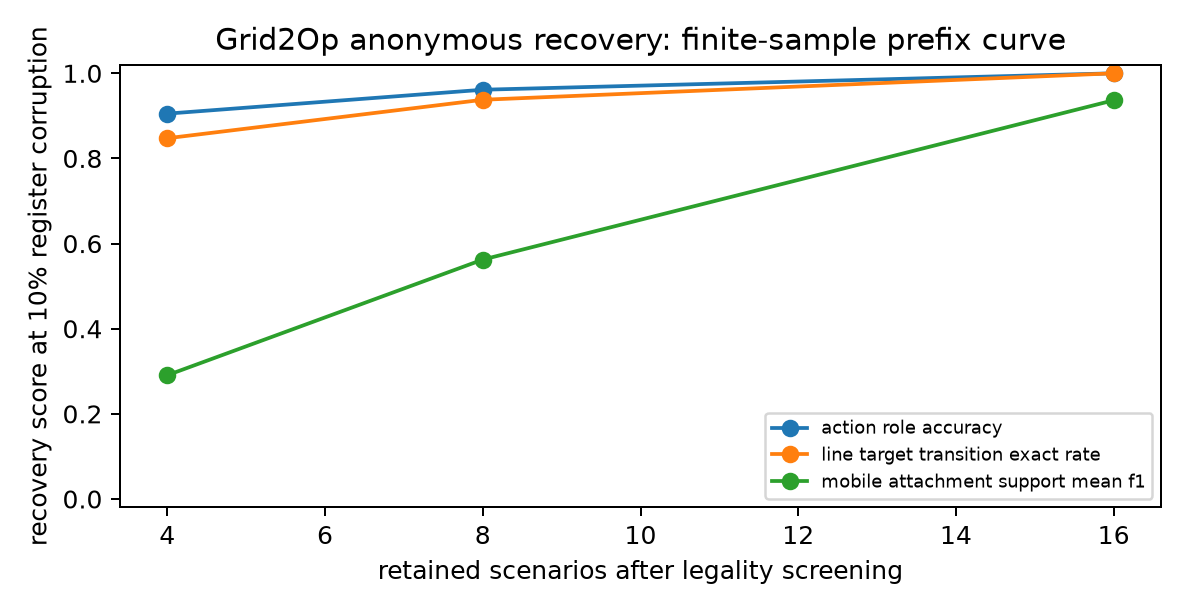}
\caption{Anonymous controlled graph-language recovery on Grid2Op's bundled
118-substation development case.  At 10\% independent register corruption,
attachment-support recovery improves sharply with repeated scenarios; the figure
does not claim operational-grid chronics or a Maxwell ontology.}
\label{fig:grid2op}
\end{figure*}

Grid2Op~\cite{donnot2020grid2op} supplies a distinct, addressable graph-intervention
contract.  On its bundled 118-substation development case, we concatenate 186
line-availability and 533 attachment-state registers, then independently permute
all coordinates and 285 retained intervention-family identities.  The recovery
program receives only anonymous before/after registers and simulator legality flags;
line/substation names, the register-sector boundary, graph, and action targets are
withheld until scoring.  To avoid enumerating exponentially many local partitions,
we directly construct one maximal two-bus split per substation.  It selects 177
legal line-disconnect and 108 legal substation-reconfiguration families from 304
proposed operations, and retains the 19 permanently invalid commands as a certificate
rather than a successful response.  The direct action library takes 3.5 ms to
construct; the 16-scenario simulator collection takes 286 s, and each recovery
pass takes at most 0.14 s.

Across 16 scenarios, zero-noise role, line-target, and mobile-attachment support
recovery are all $1.0$.  Under independent 10\% register corruption they are
$0.996$, $0.994$, and $0.939$, respectively.  The finite-sample 10\%-noise
attachment-support F1 is $0.291$, $0.562$, and $0.937$ for 4, 8, and 16 retained
scenarios.  At 30\% corruption it falls to $0.034$, consistent with the analytic
majority-signature cutoff $p<1-\sqrt{1/2}\simeq0.293$.  The protocol also certifies
134 attachment registers that never move under retained legal operations.  This is
evidence for anonymous controlled \emph{graph-language} recovery, not recovery of
a dual complex, Maxwell form degree, or operational-grid chronics.

We additionally run the downloaded public \texttt{l2rpn\_case14\_sandbox}
chronics directly, rather than a bundled Grid2Op development fixture.  Across 128
chronological scenarios, the anonymous record has 20 line-availability registers,
57 attachment-state registers, and 32 retained intervention families (19 line
disconnects and 13 substation reconfigurations).  At zero and 10\% independent
binary register corruption, respectively, action-role accuracy is $0.969$,
line-target accuracy is $0.947$, and mobile attachment-support F1 is $0.981$.
The 10\%-noise finite-sample curve reaches the same values by 32 scenarios and
stays there through 128.  Crucially, the output certifies 15 attachment registers
that no retained legal operation moves and one structurally ambiguous line response;
they are not relabelled as successes.  At 30\% corruption, role accuracy,
line-target accuracy, and attachment F1 fall to $0.656$, $0.368$, and $0.723$.
This establishes external-chronics evidence for the limited graph-language target,
not an operational-grid, dual-complex, or Maxwell claim.

\subsection{Benchmark and experiment-design criteria}
\label{app:benchmark-criteria}

An ontology benchmark should first declare its latent language: carrier and
incidence relation, measurement sectors, permitted interventions, refinement maps,
and target dual or constitutive structure.  It should then hide all human names,
bases, orientations, and cell correspondences.  Supplying a graph, form degree, or
channel role changes the inverse problem from language discovery to learning within
a supplied language.

Scores must separate discrete and continuous claims.  Carrier scoring requires a
poset/rank match modulo allowed automorphisms; transport scoring requires the correct
extensive/intensive class; dual scoring requires the full boundary-intertwining
sequence rather than one fitted cross-map.  Continuous constitutive and trajectory
residuals follow those discrete tests, with design rank and held-out split stated.
Controls retain their native objective and input contract: equation discovery may
receive oracle variables and operators, and action learners receive matched pairs.
A comparison should record which oracle objects are supplied, each native success
metric, and whether the interface returns the proposed ontology object.

Noisy benchmarks should report both a failure boundary and its repair cost.  The
one-shot 0.3\% failure, 1\% carrier-support failure, condition-number curve,
four-read 0.3\% repair, and 64-read 1\% repair diagnose different stages.  A single
mean would hide whether the required remedy is another sample, repeated readout, a
more separating fingerprint, or a new intervention.

The theory also allocates a finite measurement budget.  Carrier observations need
support and spectral margin; refinement should include unequal branching; geometry
probes should maximize the commutant gap; and excitation length should reach the
Faraday/Amp\`ere design dimension.  A reusable external benchmark should retain raw
cell-boundary observations, orientation responses, intervention matrices and
targets, refinement or registration transports, source metadata, and temporal/spatial
staggering, while revealing semantics only for evaluation.  Otherwise the correct
output may be an ingress diagnostic and a list of unbroken equivalences rather than
an invented complete ontology.

\end{document}